\documentclass{article}

\usepackage{iclr2027_conference,times}
\iclrfinalcopy 

\usepackage{pdflscape} 
\usepackage[utf8]{inputenc}
\usepackage[T1]{fontenc}
\usepackage{hyperref}
\usepackage{url}
\usepackage{float}
\usepackage{booktabs}
\usepackage{amsfonts}
\usepackage{amsmath}
\usepackage{nicefrac}
\usepackage{microtype}
\usepackage{xcolor}
\usepackage{graphicx}
\usepackage{subcaption}
\usepackage{multirow}
\usepackage{amsthm}
\usepackage{dsfont}
\usepackage{longtable}
\usepackage{enumitem}
\usepackage{array}
\usepackage[normalem]{ulem}

\newcommand{\querycode}[1]{\colorbox{gray!12}{\textcolor{red!75!black}{\texttt{#1}}}}

\usepackage{xspace}
\newcommand{\model}{EveryQuery\xspace}

\title{\model: A Promptable Foundation Model for Clinical Prediction Tasks over EHRs}

\author{%
  Payal Chandak\thanks{Equal contribution} \\ Harvard-MIT HST
  \And Gregory Kondas\footnotemark[1] \\ Columbia University
  \And Liat Antwarg Friedman \\  Harvard Medical School
  \And Isaac Kohane \\ Harvard Medical School
  \And Matthew McDermott \\ Columbia University  \\ \texttt{mattmcdermott8@gmail.com}
}
\begin{document}

\maketitle

\begin{abstract}
Autoregressive foundation models for electronic health records (EHRs) can make predictions for diverse clinical tasks without finetuning~\citep{Waxler2025-pg,Renc2024-el,Renc2025-cw}. These methods work by generating possible ``synthetic futures'' for patients, and then inferring downstream predictions over these simulated trajectories. Despite their strong capabilities, these methods have several flaws. (1) Inference is extremely computationally expensive, as each prediction requires simulating many trajectories, each with many observations. (2) Their performance is noisy due to the variance inherent to simulation, which in particular reduces efficacy on rare events, which are often of high importance clinically. (3) They are not promptable, as their only input is the patient's medical history, requiring indirect aggregation at inference time. 
We introduce \model, a promptable foundation model for structured EHR data. Building on the insight that clinical prediction tasks can be written in a structured query language, \model takes a query as a prompt alongside the patient's history and estimates the outcome directly. This allows \model to make predictions across diverse clinical tasks, such as classification and survival modeling, without simulation or finetuning.
Across three EHR datasets, \model outperforms a competitive autoregressive baseline on clinically relevant classification and time-to-event tasks, with mean AUROC gains of 14.3\% and 12.7\% on MIMIC-IV and a large academic medical center and comparable performance on the smaller NWICU dataset. Its advantage is largest for rare outcomes, and it requires $\sim580\times$ less inference compute.

\end{abstract}

\section{Introduction}
\label{sec:intro}

Foundation models for structured electronic health record (EHR) data are an emerging new paradigm in clinical artificial intelligence (AI). Much like in other AI domains, foundation models for EHR data are appealing because they can make predictions across a diverse range of clinically meaningful tasks, such as classification or survival analysis.
Currently, the only models capable of making predictions for diverse tasks at inference without finetuning are autoregressive (AR), generative models. Across numerous datasets, AR models show strong performance in estimating diverse clinical outcomes~\citep{Waxler2025-pg, Renc2024-el, Renc2025-cw, Rajamohan2025-xu, Shmatko2025-ao, McDermottUnknown-pi}.

Similar to large language models, autoregressive models are trained with a next-token prediction objective. At each training step, they perform a supervised multiclass classification problem over the model's vocabulary to predict the next medical event. Operationally, they split a patient's record into a past and future, and use the patient's past to predict which code comes first in their future. Unlike LLMs, however, which can answer novel questions directly at inference time through task prompting, AR models make predictions for specific tasks through simulation-based inference. Given as input a patient's medical history, they generate possible ``synthetic futures'' for the patient. The user must then extract a label for the task of interest from each synthetic future and estimate the probability of the outcome from how often it occurs across multiple simulated trajectories (Figure~\ref{fig:overview}a).

While AR models have reported strong performance across many clinical prediction tasks ~\citep{Waxler2025-pg, Renc2024-el, Renc2025-cw, Rajamohan2025-xu, Shmatko2025-ao, McDermottUnknown-pi}, this design has three critical problems.
First, simulation is extremely expensive. To make a single prediction, an AR model must simulate many trajectories, each of which requires generating many medical observations. As a result, \emph{inference alone} over even small datasets can take thousands of GPU hours to complete~\citep{Solo2026-pp}.
%
Second, these models are statistically noisy, and in particular struggle to predict rare outcomes. Concretely, autoregressive models estimate event probabilities by counting occurrences across $K$ sampled trajectories. For rare events with probability $p\ll1$, the expected
number of trajectories containing the event is $Kp$, which may be very small even for large $K$. As a result, sample-driven estimates of the probability of these outcomes occurring have high variance and low sensitivity~\citep{Waxler2025-pg,Renc2025-cw}.
%
Third, AR models are not promptable: the task of interest is not an input to the model. Predictions for a given task are therefore obtained only indirectly, by post-processing simulations, rather than by querying the model for that task directly.

Here, we address these shortcomings by leveraging the insight that clinical prediction tasks are expressible in a parametric language. Given any structured query at inference, our model, \model, takes this query as an input prompt along with the patient's historical record and estimates the corresponding future outcome. Whereas autoregressive pretraining operates on medical events, \model learns a shared structure over medical events and a query language for clinical prediction tasks. As a result, \model is promptable, and can make direct predictions for clinically meaningful tasks without simulation (Figure~\ref{fig:overview}b).
We validate our approach on three datasets, including MIMIC-IV \citep{Johnson2023-tp}, NWICU \citep{nwicu}, and a large academic medical center hereafter referred to as ``AMC'', and make the following contributions:
\begin{enumerate}[leftmargin=0.35cm]
    \item \textbf{A new pretraining approach for clinical prediction on structured EHR data}: Pretraining jointly over medical events and a query language for clinical tasks is a viable alternative to generative simulation, opening a new direction for EHR foundation models.

    \item \textbf{A foundation model with superior performance across diverse clinical tasks}: \model outperforms an autoregressive baseline on both classification and time-to-event prediction tasks. Across clinically meaningful tasks, \model achieves win rates of 100\% on MIMIC-IV, 94\% on AMC, and 66.7\% on NWICU (Wilcoxon signed-rank $p < 10^{-4}$ on MIMIC-IV and AMC; not significant on NWICU). \model demonstrates significant performance gains on rare outcomes.

    \item \textbf{Efficient and tractable inference compared to autoregressive models}: As \model can be prompted with tasks at inference, outcomes can be estimated directly from the model. \model requires 582$\times$ lower compute than AR modeling and simulation-based inference.
\end{enumerate}

\section{Background}
\label{sec:background}
\paragraph{Structured clinical data.}
A patient's structured electronic health record (EHR) is a longitudinal sequence of discrete clinical events sourced from a particular medical institution. Each observation can be represented as a triple of a timestamp, a categorical medical ``code'' (e.g., a diagnostic code or laboratory test), and an optional numeric value~\citep{Arnrich2024-uy}. Increasingly, however, models discretize all aspects of these observations into a sequence of purely categorical ``tokens''. In this scheme, continuous numeric values (e.g., laboratory test results) and time deltas between observations are represented through quantized categorical tokens. As such, the entire EHR dataset is represented as ordinal sequences over a unified vocabulary of tokens which we will denote $\mathcal{V}$~\citep{Renc2024-el,wornow2025context,Guo2026-ec,Lee2026-ky}. We write a patient's tokenized record as $\mathbf{x} = (x_1, \ldots, x_L)$ with $x_i \in \mathcal{V}$, where elapsed time $t_j$ at any position $j$ is recovered by accumulating the time-delta tokens up to $j$.  Because $\mathcal{V}$ is highly specific to the medical institution from which the EHR dataset is derived, foundation models trained over these token sequences are generally trained and evaluated within a single dataset and cannot be transferred to a new dataset without additional work to harmonize vocabularies. Most existing EHR foundation models follow this practice~\citep{Waxler2025-pg,Renc2024-el,Renc2025-cw}.

\paragraph{Prediction tasks on EHR data.}
EHR data is used to predict diverse tasks of significant clinical interest, such as whether a patient will be readmitted to the hospital within 30 days~\citep{tang2023predicting}, whether a patient is at imminent risk of mortality~\citep{Pang2025-md,harutyunyan2019multitask}, or how long it will take until the patient first develops a disease~\citep{Shmatko2025-ao, Steinberg2023-yy}. It is well established that clinical prediction tasks can be expressed in a deterministic, parametric form, through, e.g., SQL queries or ACES~\citep{Xu2024-qr} configuration files. Such task definitions typically share a common structure: an outcome of interest, a prediction window, and optionally events that condition on or bound that window (e.g., predicting in-hospital mortality only for patients admitted to the ICU, and only until discharge). This structure supports many forms of prediction, including binary classification, joint prediction of several outcomes, and time-to-event estimation.

\begin{figure}[t]
  \centering
  \includegraphics[width=\textwidth]{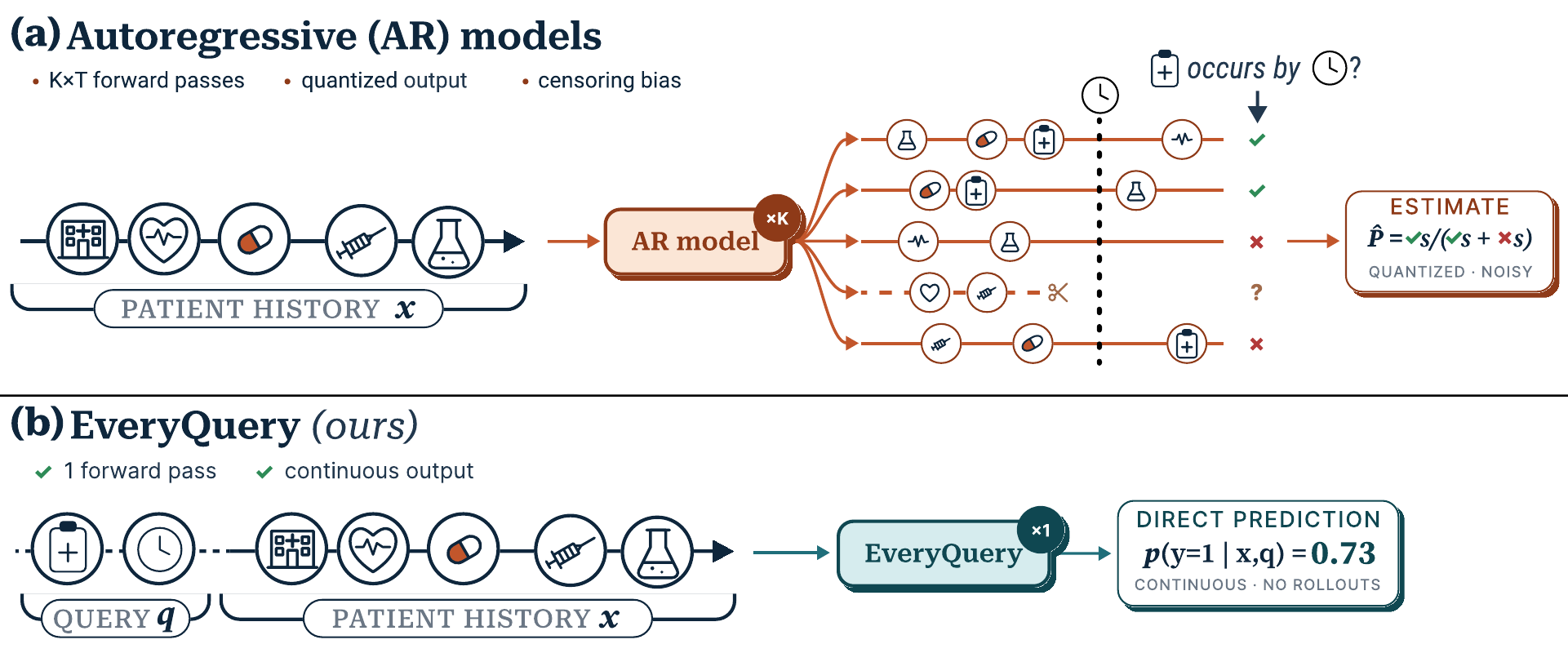}
\caption{\textbf{Overview of \model.} Both models answer a query $\mathbf{q}$ during inference (\S\ref{sec:query_language}), estimating whether a code occurs before a given time, for a patient history $\mathbf{x}$. \textbf{(a)} An autoregressive model generates $K$ future trajectories of up to $T$ events each, and estimates the probability as the fraction of trajectories in which the code occurs. This yields quantized, high-variance estimates. \textbf{(b)} \model takes $\mathbf{q}$ as input alongside $\mathbf{x}$ and predicts $p(y = 1 \mid \mathbf{x}, \mathbf{q})$ directly.}
  \label{fig:overview}
\end{figure}

\paragraph{Autoregressive EHR foundation models.}
AR models ingest a patient's EHR data as a tokenized sequence and predict the next token, i.e., they model $p_\theta(x_i \mid x_{<i})$ through iterative multiclass classification over $\mathcal{V}$ at each position. This pretraining regime makes AR models capable of generating arbitrary token sequences, or ``simulated futures'' for the patient. Under a simple Monte Carlo approach, predictions on clinical tasks are made by extracting the target label from each synthetic future and aggregating these labels into an empirical distribution (Figure~\ref{fig:overview}a)~\citep{Waxler2025-pg, Renc2024-el,Renc2025-cw}. Advanced methods, such as importance sampling~\citep{Solo2026-pp}, improve efficiency, but prediction still requires hundreds to thousands of inference passes.

\paragraph{Task-specific and representation-learning models.} Several other families of models are trained on structured EHR data, but none can make predictions across a diverse set of clinical tasks arbitrarily defined at inference time. Supervised models such as XGBoost~\citep{chen2016} are trained for a single, pre-specified task and are unable to generalize to new tasks. Multitask models such as MOTOR~\citep{Steinberg2023-yy} and M3H~\citep{Bertsimas2024-xu} predict a  large, but fixed and incomplete set of tasks through dedicated output heads chosen before training, and are unable to adapt to tasks outside this set. Representation-learning models that pretrain bidirectional or encoder-decoder representations on EHR data \citep{Rasmy2021-jv, Pang2021-qq, Wornow2023-xm, Yang2023-nc, clmbr} produce patient embeddings that require fine-tuning or linear probing for each new task, and therefore cannot make predictions for tasks specified at inference without task-specific labeled data. Among existing approaches, only AR models can address diverse clinical prediction tasks, from binary classification to time-to-event estimation, with a single pretrained model at inference time.

\section{The \model Method}
\label{sec:methods}

Driven by the insight that clinical prediction tasks can be expressed in a simple, structured form, we propose an alternative to autoregressive pretraining that sidesteps trajectory sampling entirely. \model learns representations of the patient history $\mathbf{x}$ conditioned on a query task $\mathbf{q}$ by taking both as input and predicting the answer directly, $p_\theta(y \mid \mathbf{x}, \mathbf{q})$ (Figure~\ref{fig:overview}b). 
During pretraining, we construct pairs of patient histories and synthetic tasks, where tasks are randomly sampled from a continuous distribution over a query language. This query language distribution is specified during pretraining in a manner that is completely agnostic to inference tasks. \model has the same ``API'' as autoregressive models, where it is pre-trained over an unlabeled EHR dataset in  a task-agnostic manner, and can then answer diverse prediction tasks at inference time.

\subsection{Query Language}
\label{sec:query_language}

\paragraph{Structured queries.}
\model leverages the natural parametric structure of clinical prediction tasks~\citep{Xu2024-qr}. We define the basic unit of our query language as a \emph{structured query} $q_i = (c_i, w_i)$, which asks whether a clinical event $c_i \in \mathcal{V}$ occurs within a future window $w_i = (s_i, e_i)$. Each window boundary is either a duration in $\mathbb{R}_{+}$ or a clinical event in $\mathcal{V}$, which resolves to the time of that event's next occurrence. A duration-specified start is defined relative to the prediction time $t_L$, whereas a duration-specified end is defined relative to the resolved window start. With $\tau(s_i)$ and $\tau(e_i)$ denoting the resolved start and end times, the answer for patient $\mathbf{x}$ is $\mathds{1}_{q_i}(\mathbf{x}) = \mathds{1}[\exists\, j > L : x_j = c_i,\ \tau(s_i) < t_j \le \tau(e_i)]$. Event-bounded windows allow structured queries to follow the course of care rather than fixed durations. Table~\ref{tab:query_examples} lists examples.

\begin{table}[t]
\centering
\footnotesize
\setlength{\tabcolsep}{3pt}
\setlength{\fboxsep}{1.5pt}
\caption{\textbf{Example prompts in the query language.}}
\label{tab:query_examples}
\begin{tabular}{lll}
\toprule
\textbf{Task} & \textbf{Conditional Clauses $(q_i, a_i)$} & \textbf{Target $q_n$} \\
\midrule
In-hospital mortality & -- & \querycode{(death, (0, discharge))} \\
30-day readmission & -- & \querycode{(admission, (discharge, 30d))} \\
Post-discharge mortality & \querycode{((death, (0, discharge)), 0)} & \querycode{(death, (discharge, 90d))} \\
\bottomrule
\end{tabular}
\end{table}

\paragraph{Conditional clauses.}
A structured query can be extended with \emph{conditional clauses}, which enable complex questions about future events, in which a target structured query $q_n$ is asked given that other structured queries $q_i$ do ($a_i = 1$) or do not ($a_i = 0$) occur in the patient's future. We formalize a prompt input $\mathbf{q} = ((q_0, a_0), \ldots, (q_{n-1}, a_{n-1}), q_n)$ as a target structured query $q_n$ asked conditional on the clauses $(q_i, a_i)_{i<n}$, where each clause specifies the observed outcome $a_i \in \{0,1\}$ of a preceding structured query $q_i$.  The answer to $\mathbf{q}$ is defined only for patients whose futures satisfy every condition,
$y(\mathbf{x}, \mathbf{q}) = \mathds{1}_{q_n}(\mathbf{x}) \quad \text{if } \mathds{1}_{q_i}(\mathbf{x}) = a_i \;\; \forall i < n,$
and \model estimates $p_\theta(y = 1 \mid \mathbf{x}, \mathbf{q}) \approx P\big(\mathds{1}_{q_n}(\mathbf{x}) = 1 \mid \mathbf{x},\, \mathds{1}_{q_i}(\mathbf{x}) = a_i \;\forall i < n\big)$. With $n = 0$, the prompt is a single structured query with no conditional clauses.


\paragraph{Clinical ontology.}
Clinical questions are often asked about groups of codes rather than a single code. For example, a clinician may want to know whether a patient will be discharged at all, not whether they will be discharged home or to a skilled nursing facility, which are separate codes in the vocabulary. \model supports such questions by arranging the codes of the dataset vocabulary into a hierarchy, in which a parent code such as \querycode{discharge} stands for all of its more specific descendants. Every parent code is added to the vocabulary $\mathcal{V}$ with its own embedding, and a structured query about a parent code is answered positively if any of its descendants occurs in the window.

\subsection{Model}
\label{sec:model}

\paragraph{Input sequence.}
Given a patient history $\mathbf{x}$ up to the prediction time $t_L$ and a prompt $\mathbf{q}$ with conditional clauses $(q_i, a_i)_{i<n}$ and target structured query $q_n$ (\S\ref{sec:query_language}), \model processes a single sequence
\[ [\,x_1, \ldots, x_L,\; w_0, c_0, a_0,\; \ldots,\; w_{n-1}, c_{n-1}, a_{n-1},\; w_n\,], \] 
in which each conditional clause contributes a window token $w_i$, a code token $c_i$ and an answer token $a_i$, and the target structured query $q_n = (c_n, w_n)$ contributes only its window token $w_n$. Patient history is truncated to the most recent 256 events. \model uses a Llama-style decoder-only transformer~\citep{touvron2023llama} trained from scratch, with standard causal attention: each window token $w_i$ attends to the full patient history, to every earlier conditional clause, and to itself, but never to the answer for its own window.

\paragraph{Time and position.}
\model encodes time through its position embeddings rather than via time-delta tokens. Time-delta tokens are removed from the input after their durations are accumulated, which shortens sequences so that more clinical events fit in the context window. Each preserved event's rotary position embedding is computed from its elapsed time in hours rather than its index in the sequence. All query tokens are assigned the prediction time $t_L$, anchoring them to the end of the patient context. A nonzero window start is expressed through its window token rather than position.

\paragraph{Token embeddings.}
Every input token is embedded as the sum of two learned vectors: a \emph{content} vector, which encodes what the token says, and a \emph{token-type} vector denoted $\mathbf{t}_{\text{type}}$, which encodes what kind of token it is. There are four token types: patient event, window, clause code and clause answer. Query tokens additionally receive a \emph{block} vector $\mathbf{b}_i$, which identifies the structured query $q_i$ they belong to. The resulting embeddings are:
\begin{center}
\begin{tabular}{lll}
\toprule
\textbf{Token} & \textbf{Embedding} & \textbf{Content vector} \\
\midrule
Patient event $x_j$ & $E[x_j] + \mathbf{t}_{\text{patient}}$ & code embedding \\
Window $w_i$ & $\mathbf{u}(s_i) + \mathbf{v}(e_i) + \mathbf{t}_{\text{window}} + \mathbf{b}_i$ & start and end embeddings \\
Clause code $c_i$ & $E[c_i] + \mathbf{t}_{\text{code}} + \mathbf{b}_i$ & code embedding \\
Clause answer $a_i$ & $A[a_i] + \mathbf{t}_{\text{answer}} + \mathbf{b}_i$ & answer embedding \\
\bottomrule
\end{tabular}
\end{center}
All codes, whether patient events, clause codes or window boundaries, are embedded with a single shared table $E$. Numeric values such as laboratory results are binned into the code during preprocessing. Answers and durations have embeddings of their own: answers use a learned two-entry table $A$, and durations use the MLPs described below. When the clinical ontology (\S\ref{sec:query_language}) is used, each row of $E$ is replaced by a normalized average of that code's embedding and its ancestors' embeddings, with weights that decay with distance, so that related codes, including rare ones, share information.

\paragraph{Window boundaries.}
The start embedding $\mathbf{u}(s_i)$ and end embedding $\mathbf{v}(e_i)$ depend on how each boundary is specified. An event boundary is embedded as the event's code embedding plus a learned marker to distinguish start from end, $\mathbf{u}(s_i) = E[s_i] + \mathbf{m}_{\text{start}}$ or $\mathbf{v}(e_i) = E[e_i] + \mathbf{m}_{\text{end}}$. Temporal start and end boundaries are converted to years and processed by separate MLPs.

\paragraph{Readout.}
The target structured query $q_n = (c_n, w_n)$ asks whether code $c_n$ occurs in window $w_n$. The window
$w_n$ is given to the model as an input token, but the code $c_n$ is not. Instead, the model reads out
its answer by comparing its output at $w_n$, denoted $\mathbf{h}_n$, with the embedding of $c_n$:
\[
p_\theta(y = 1 \mid \mathbf{x}, \mathbf{q}) = \sigma\big(\mathbf{h}_n^\top E[c_n] + \beta_{c_n}\big),
\]
where $E$ is the same code embedding table used for the inputs and $\beta_{c}$ is a learned per-code
bias.

\subsection{Pretraining}
\label{sec:pretraining}

\paragraph{Objective.}
\label{sec:pretraining_objective}
The pretraining objective is to predict the answer to every structured query in a training prompt. Because $\mathbf{h}_n$ does not depend on $c_n$, the readout above can score every code $c \in \mathcal{V}$ at once. We apply it at every window token $w_i$, not only the target, so that each window in a training prompt supervises $|\mathcal{V}|$ structured queries that share its window and preceding clauses, or $(n+1)|\mathcal{V}|$ structured queries per prompt. The model is trained with binary cross-entropy averaged over all windows and codes, with the clause answers $a_i$ teacher-forced from the patient's true future. \model therefore learns $p_\theta\big(\mathds{1}_{(c, w_i)}(\mathbf{x}) = 1 \mid \mathbf{x}, w_0, \ldots, w_i, (c_j, a_j)_{j<i}\big)$ for every code $c$ and every window $i$. Whereas autoregressive pretraining performs multiclass classification over $\mathcal{V}$, predicting which one code comes next, \model solves $|\mathcal{V}|$ simultaneous binary classification problems at each window. Supervising this many queries per prompt reduces the variance of the gradient.

\paragraph{Training.}
We construct 5 million training contexts. For each, we draw a patient uniformly at random, then sample a prediction time uniformly after excluding times with fewer than 50 distinct preceding timestamps, and sample a query with five windows. Each window starts at the prediction time (40\%), after a delay drawn log-uniformly from 1 to 180 days (40\%), or at an event (20\%), and ends either at an event (50\%) or after a duration drawn log-uniformly from 1 to 1,826 days (50\%). Boundary events and clause codes are drawn uniformly over all codes, including ontology ancestors, independently of the patient's data. Further optimization details are in Appendix~\ref{app:hyperparameters}.

\section{Experiments}
\label{sec:experiments}
 
\paragraph{Datasets.} We train and evaluate on three EHR datasets, each preprocessed into the MEDS format \citep{mcdermott2026}: MIMIC-IV (v2.2) \citep{Johnson2023-tp}, NWICU \citep{nwicu}, and a private EHR dataset from a large academic medical center, hereafter AMC. The three datasets vary substantially in population, care setting, and scale, with population sizes of 240K, 24K, and 1.9M patients after preprocessing. Data splits are described in Appendix~\ref{app:datasets}. All results are reported on the held-out test split of each dataset. Following convention, we train one \model and one autoregressive baseline on each dataset.
 
\paragraph{Baseline.} 
We compare against MEDS-EIC-AR \citep{McDermottUnknown-pi}\footnote{\url{https://github.com/mmcdermott/MEDS_EIC_AR}}, an open-source ``everything-is-code'' autoregressive model that fuses each clinical code with its decile-binned value into a single token, a design supported by recent tokenization benchmarks for structured EHR foundation models \citep{Lee2026-ky, Guo2026-ec}. MEDS-EIC-AR is one of the few released AR foundation models that can be trained on any MEDS dataset, and it has been shown to be effective for pretraining, fine-tuning and representation learning~\citep{pollet2026rmethodprogress}. The model has 102M parameters, uses a max sequence length of 1024 tokens, of which at most 256 are reserved for the input patient context and the remainder for autoregressive generation via next-token prediction. Both \model and MEDS-EIC-AR therefore operate on the same effective patient history context of 256 tokens at inference time, though MEDS-EIC-AR benefits from longer contexts during training. At inference, we sample $K = 20$ future trajectories per patient, as is typical \citep{Renc2025-cw}, and estimate each task's probability from the labels extracted from these trajectories.

\paragraph{Evaluation.} 
We evaluate on diverse classification tasks, each anchored at a clinically significant prediction time and optionally defined over a cohort of interest. The 38 and 27 tasks on the MIMIC-IV and NWICU datasets are anchored at 24 hours after ICU admission. The 100 tasks on the AMC dataset in the longitudinal care setting are anchored at outpatient visits, and some are defined over type~2 diabetes and heart failure cohorts. Tasks cover a wide range of clinical events described in Figure~\ref{fig:tasks}. Where appropriate, tasks are conditioned on the patient being neither censored nor deceased within the window, and some add other task-specific clauses. 

To further investigate tasks that are not directly expressible in the query language, we evaluate whether \model can recover survival curves for events of interest, and report comparisons to Kaplan--Meier estimates. Given the time $T$ to the first occurrence of a target code $c$, with records censored at their last event, the survival function is $S(t) = P(T > t)$. \model estimates the survival function at each time $t$ from three structured queries: the probability that $c$ occurs by $t$, $F(t)$; the probability that the record ends by $t$, $Q(t)$; and the probability that $c$ occurs by $t$ given that the record remains open, $U(t)$. Together these give the probability of being event-free and still at risk, $R(t) = (1 - Q(t))(1 - U(t))$, and hence the hazard $\lambda(t) = F'(t)/R(t)$, the same quantity that Kaplan--Meier estimates. The survival curve then follows as $S(t) = \exp\big(-\int_0^t \lambda(s)\,ds\big)$. The autoregressive baseline instead pools its $K$ trajectories into a per-patient Kaplan--Meier curve. We report Harrell's C, and list target codes and prediction horizons for each dataset in Appendix~\ref{app:tte}. 
 
\section{Results}
\label{sec:results}

\begin{figure}[t]
  \centering
  \includegraphics[width=\textwidth]{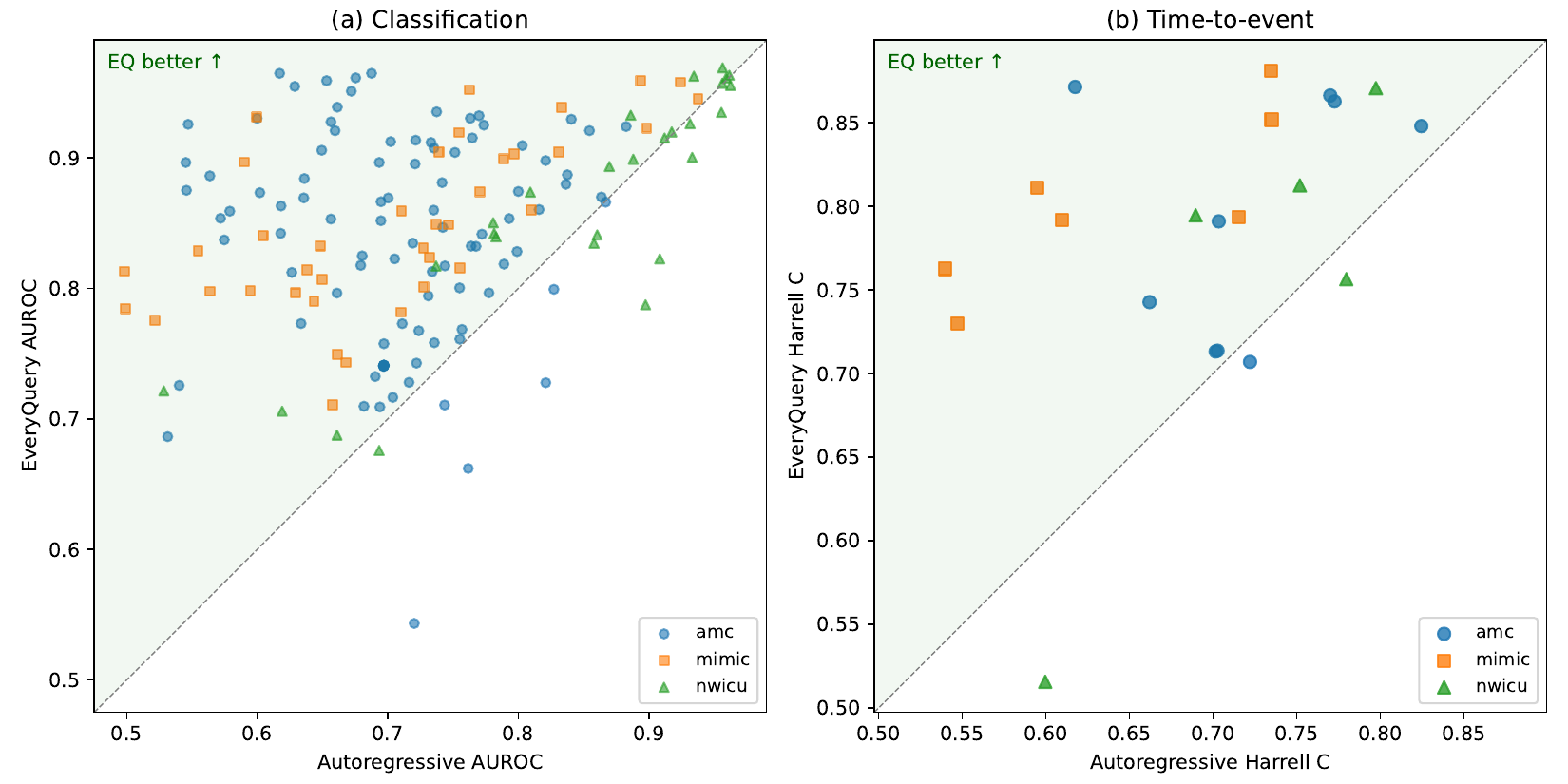}
\caption{\textbf{\model outperforms the autoregressive baseline across clinical prediction tasks.} Each point is a task, comparing \model with the AR baseline on AUROC for classification tasks (a) and Harrell's C for time-to-event tasks (b). Points above the diagonal favor \model.}
  \label{fig:performance}
\end{figure}

\subsection{\model outperforms AR baseline at diverse clinical prediction tasks}
\label{sec:performance}

\model outperforms the autoregressive baseline on both classification and time-to-event tasks, with large gains on MIMIC-IV and AMC and comparable performance on NWICU (Figure~\ref{fig:performance}).

\paragraph{Classification.} 

On the clinically relevant classification tasks, \model achieves a higher AUROC than the autoregressive baseline on 100\% of tasks on MIMIC-IV, 66.7\% on NWICU and 94\% on AMC (Figure~\ref{fig:performance}a). The win rates on MIMIC-IV and AMC are significantly above chance (Wilcoxon signed-rank $p<10^{-4}$). Averaged over tasks, \model reaches AUROCs of 84.9\%, 86.6\% and 83.6\% on MIMIC-IV, NWICU and AMC. Its mean AUROC gain over the autoregressive baseline is $+14.3\%$ (95\% CI $[11.7\%, 17.0\%]$) on MIMIC-IV and $+12.7\%$ ($[10.5\%, 14.8\%]$) on AMC. On NWICU, the mean gain is smaller and not significant ($+1.6\%$, $[-0.4\%, 3.8\%]$), consistent with the lower win rate on this dataset. Per-task AUROCs are listed in Appendix~\ref{app:class}.

\paragraph{Time-to-event.} Although time-to-event tasks are not directly expressible in the query language, EveryQuery still outperforms the autoregressive baseline. \model reaches a mean Harrell's C of 80.3\%, 75.0\% and 79.1\% on MIMIC-IV, NWICU and AMC. \model achieves mean per-task win margins over the autoregressive baseline of $+16.4\%$ (95\% CI $[11.5\%, 21.3\%]$), $+2.6\%$ ($[-7.0\%, 12.3\%]$) and $+7.1\%$ ($[0.9\%, 13.3\%]$) (Figure~\ref{fig:performance}b). The margins on MIMIC-IV and AMC are significant in Wilcoxon signed-rank test, while NWICU is again comparable. Survival curves for all tasks are shown in Appendix~\ref{app:tte}.

\subsection{\model is strongest when making predictions for rare events}
\label{sec:rare_events}

\begin{figure}[t]
  \centering
  \includegraphics[width=1\textwidth]{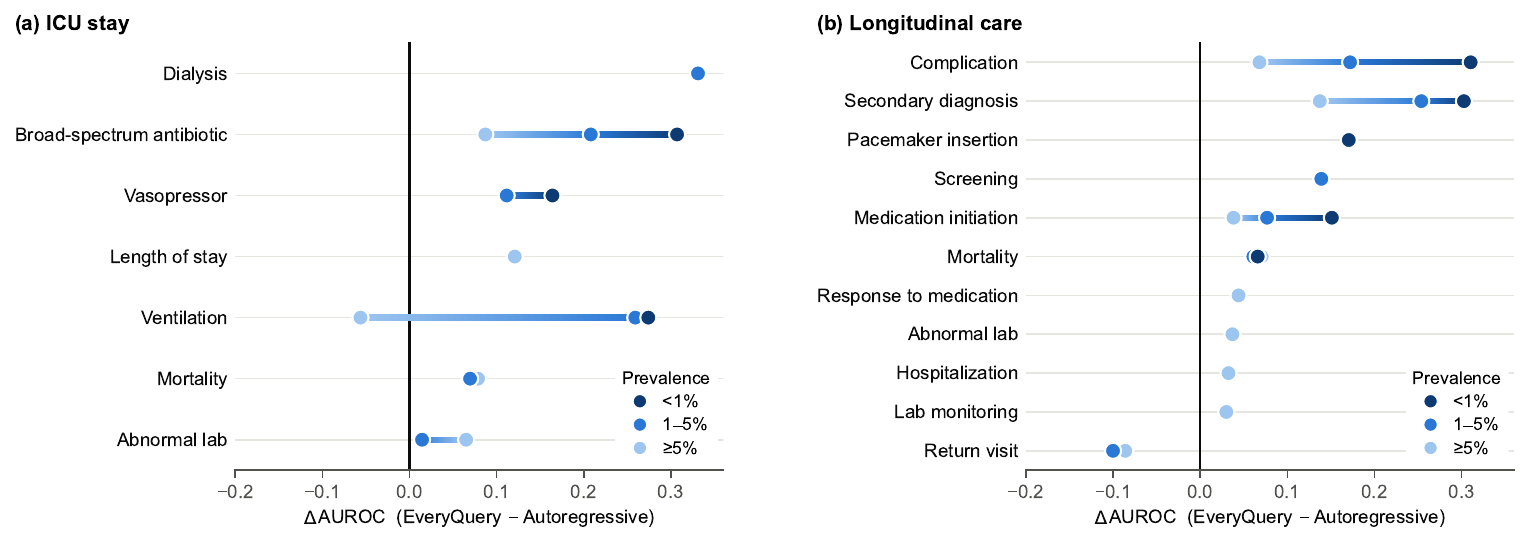}
\caption{\textbf{\model's advantage over the autoregressive baseline is largest for rare events.} Each row is a category of clinical prediction tasks, and each dot shows the mean $\Delta$AUROC for the tasks in that category grouped by prevalence. ICU tasks \textbf{(a)} are evaluated on MIMIC-IV and NWICU; and longitudinal care tasks \textbf{(b)} on AMC. EveryQuery's gains grow as outcomes become rarer.}
  \label{fig:tasks}
\end{figure}

\model performs best on rare events, which are often the outcomes of greatest clinical interest and the hardest for simulation-based models to estimate (Figure~\ref{fig:tasks}). Across classification tasks on MIMIC-IV, NWICU and AMC, \model's AUROC is negatively correlated with task prevalence on all three datasets (Spearman $\rho = -0.40$, $p = 0.014$ on MIMIC-IV; $\rho = -0.59$, $p = 0.001$ on NWICU; $\rho = -0.38$, $p = 10^{-4}$ on AMC), so \model has better discrimination on rarer outcomes.

As a result, \model's advantage over the autoregressive baseline grows as events become rarer. The win margin ($\Delta$AUC) is negatively correlated with prevalence on MIMIC-IV ($\rho = -0.35$, $p = 0.03$) and AMC ($\rho = -0.61$, $p = 10^{-11}$). The autoregressive baseline does not share \model's advantage on rare events in these datasets: its AUROC is uncorrelated with prevalence on MIMIC-IV ($\rho = +0.11$, $p = 0.52$) and positively correlated with prevalence on AMC ($\rho = +0.45$, $p = 2\times10^{-6}$), meaning it performs worse on rarer events. On NWICU, the correlation between win margin and prevalence is not significant ($\rho = +0.23$, $p = 0.24$). 

\subsection{\model is more efficient at inference than autoregressive modeling}
\label{sec:efficiency}
 
\model requires a single deterministic forward pass per (patient, query) pair. The autoregressive baseline requires generating $K=20$ full future trajectories, each of which has hundreds of events on average, per patient. To compare inference efficiency independently of hardware, we measure the number of floating point operations (FLOPs) required per  (patient, query) for each model. \model requires only 7.76 giga FLOPs per query, compared with an average of 4.52 tera FLOPs for the optimized autoregressive baseline, corresponding to approximately 582$\times$ lower inference compute required for EveryQuery. Notably, this comparison already assumes a heavily optimized autoregressive inference implementation that reuses the prefilled patient context across all ($K=20$) generated trajectories. Without this level of optimization, using na\"ive autoregressive generation directly, \model achieved approximately 5,500 to 10,300 times faster queries-per-second during inference.

\subsection{\model's predictions respect temporal and hierarchical structure}

\model is not constrained to give consistent answers to related queries, but the structure of the query language implies that some of its answers should agree. Consider the prediction horizon, where the probability that a code occurs should be monotonically increasing in the length of the window given that the windows have the same start time. We therefore checked whether \model's predicted probabilities are monotonically increasing as the horizon grows. With a tolerance of $10^{-3}$ in predicted probability, \model's predictions are monotonically increasing across all consecutive horizons for 98.5\% of (patient, code) pairs on MIMIC-IV and 87.5\% on NWICU. \model thus learns a relatively coherent notion of time without any architectural constraint.

The clinical ontology implies a similar constraint across codes. An ancestor code occurs whenever any of its descendants is observed (\S\ref{sec:query_language}), so for the same patient and window, the probability of the ancestor should be at least as high as the probability of each of its descendants. We tested this for five ontology subtrees in MIMIC-IV and NWICU by comparing, for each patient and window, the predicted probability of the ancestor with that of each of its descendants. The ancestor received a probability at least as high in 98.4\% of these comparisons on MIMIC-IV and 99.6\% on NWICU. \model therefore also learns the hierarchical structure of the ontology.

\subsection{\model's performance is robust across the query language}
\label{sec:random-tasks}

Since the clinically relevant tasks above were manually curated, we assess whether EveryQuery's performance remains robust across the query language. We repeat the evaluations above on randomly sampled tasks, following prior work showing that relative performance on randomly sampled clinical prediction tasks tracks relative performance on expert-authored tasks~\citep{pollet2026rmethodprogress}. For each dataset, we sample 1,000 prompts and reject tasks with fewer than 40 positive cases following~\citep{pollet2026rmethodprogress}.

\model remains robust to choice of tasks. \model achieves a higher AUROC than the autoregressive baseline on 100\% of tasks on MIMIC-IV, 95\% on NWICU and 100\% on AMC, and its macro-averaged AUROC is significantly higher on every dataset (Wilcoxon signed-rank $p<10^{-4}$). As in \S\ref{sec:rare_events}, \model's win margin is also significantly negatively correlated with task prevalence on all three datasets with Spearman $\rho=-0.50$, $-0.40$, and $-0.59$ for MIMIC-IV, NWICU, and AMC, respectively all with $p<10^{-4}$ (Appendix Figure~\ref{fig:random-task-prev}). \model's advantages are therefore not specific to the tasks we selected, but hold across the query language.

\section{Discussion and Conclusion}
\label{sec:discussion}
Our evaluations of \model in Section~\ref{sec:results} demonstrate that learning a shared structure over medical events and a query language for clinical tasks is an advantageous alternative to autoregressive generation for the development of EHR foundation models. In particular, across all three datasets we study, \model shows clear advantages in performance and computational efficiency over a compelling AR model baseline across both binary classification and survival modeling tasks. In this section, we offer further commentary on these results, including specific limitations of \model that should be investigated and future work.

\paragraph{Limitations and future work.}
\model has several key limitations that should be explored in future work. First and most critically, while the query language we study here is quite expressive and permits EveryQuery to directly model a wide variety of clinical prediction tasks of different types, it is not truly universal. A key area of future work, thus, is to explore extending \model's query language to cover all possible tasks over structured EHR data, such as by using a raw text expression as a query during pre-training. Beyond the query language specificity, \model's pre-training program also samples from uniform distributions over synthetic queries and patient contexts. In the real world, only a subset of these pairs are actually of clinical interest, and training efficiency may be able to be improved through more intelligent sampling methods leveraging ideas from curriculum learning. While we selected MEDS-EIC-AR for its many strengths, comparisons with other dataset-specific EHR foundation models, such as ETHOS~\citep{Renc2024-el}, would further strengthen our conclusions. Finally, like most foundation models over EHR data, \model can only be used on datasets sharing the same code vocabulary $\mathcal V$ as was seen during training. Methods to account for this, such as using universal, text-based encoders~\citep{genhpf}, could be explored to help this.

\paragraph{Conclusion.}
\model offers an alternative to autoregressive generation for EHR foundation models. Across diverse clinical prediction tasks, it reliably outperforms a competitive autoregressive baseline in both performance and efficiency, with the largest gains on rare events, while using far less inference compute. This approach offers a compelling new direction for EHR foundation model research and further exposes new opportunities for research in other longitudinal domains where tasks can be expressed parametrically.



\clearpage

\section*{AI Use Statement}
We used generative AI throughout, including to implement methods, help formalize the query language, perform limited experimental loop supervision, draft the paper, help make figures, summarize related literature, and format references. We did not use AI to interpret results or generate synthetic datasets. Proofs, translation, and qualitative or thematic data analysis are not applicable. We reviewed all AI-assisted work, including code, reported numbers, and citations. We take responsibility for the final content of this work, including text, claims, or artifacts produced with the aid of generative AI.

\section*{Ethics Statement}

This work uses de-identified electronic health record data. MIMIC-IV and NWICU were accessed using credentials. AMC data use an IRB-approved protocol (details withheld for anonymous review).

\section*{Reproducibility Statement}
We provide complete details about data, preprocessing, model architecture, pretraining, and optimization. \model is MEDS~\citep{mcdermott2026} compliant and can be trained on other MEDS datasets without modification. On publication, all our code will be released publicly.


\clearpage 

\ificlrfinal
\subsubsection*{Acknowledgments}
MBAM gratefully acknowledges support from a Berkowitz Postdoctoral Fellowship, of the Ivan and Francesca Berkowitz Family Living Laboratory Collaboration at Harvard Medical School and Clalit Research Institute
\fi

\bibliographystyle{iclr2027_conference}
\bibliography{ref}

\clearpage

\appendix

\section{Datasets}
\label{app:datasets}

\begin{table}[h]
\centering
\caption{Dataset statistics per split after preprocessing.}
\label{tab:datasets}
\small
\begin{tabular}{lrrrrr}
\toprule
& \multicolumn{3}{c}{Subjects} & & \\
\cmidrule(lr){2-4}
Dataset & Train & Val & Test & Total Events & Vocab \\
\midrule
MIMIC & 200{,}773     & 25{,}059  & 15{,}059 & 779{,}733{,}629     & 11{,}476 \\
AMC   & 1{,}651{,}030 & 291{,}538 & 27{,}019 & 2{,}250{,}292{,}868 & 21{,}289 \\
NWICU & 20{,}723      & 2{,}590   & 772      & 54{,}785{,}364      & 2{,}159  \\
\bottomrule
\end{tabular}
\end{table}

\section{Architecture and Training Details}
\label{app:hyperparameters}

Table~\ref{tab:hyperparameters} summarizes the training and architecture hyperparameters used for \model.

\begin{table}[h]
\centering
\caption{\textbf{\model training and architecture hyperparameters.}}
\label{tab:hyperparameters}
\small
\begin{tabular}{ll}
\toprule
\textbf{Hyperparameter} & \textbf{Value} \\
\midrule
\multicolumn{2}{l}{\emph{Optimization}} \\
Optimizer & AdamW \\
Learning rate & $5\times10^{-4}$\\
Betas & $(0.9, 0.98)$ \\
Epsilon & $10^{-8}$ \\
Weight decay & 0.05 \\
LR schedule & Cosine with warmup \\
Warmup steps & 5208 \\
Training steps & 104{,}167 \\
Wall-clock time & 135 mins \\
Batch size & 96 \\
Precision & bf-16 mixed \\
\midrule
\multicolumn{2}{l}{\emph{Architecture}} \\
Backbone & Llama decoder-only (from scratch) \\
Hidden dimension & 384 \\
Transformer layers & 6 \\
Attention heads & 6 \\
Backbone parameters & $\sim$22M \\
Max sequence length & 256 \\
MLP dropout & 0.1 \\
\midrule
\multicolumn{2}{l}{\emph{Data}} \\
Min.\ context times per prediction & 50 \\
Sequence sampling strategy & TO\_END \\
Padding side & RIGHT \\
Batch mode & SM \\
\midrule
\multicolumn{2}{l}{\emph{Infrastructure}} \\
GPU & 1$\times$ NVIDIA L40S (48\,GB) \\
\bottomrule
\end{tabular}
\end{table}




\clearpage

\section{Results on Classification Tasks}
\label{app:class}

Tables below report AUROC gaps for \model and the autoregressive baseline across all classification tasks, sorted by task prevalence. Tasks are sorted by descending prevalence, the better AUROC per task is in bold.

\providecommand{\bs}{/\allowbreak}  

\begin{longtable}{>{\raggedright\arraybackslash\ttfamily}p{0.62\linewidth} r r r}
\caption{AMC outpatient visit, general cohort ($n=5$ tasks).}\label{tab:app-amc-general}\\
\toprule
Task & Prev. & \multicolumn{2}{c}{AUROC} \\
\cmidrule(lr){3-4}
 & & EveryQuery & Autoregressive \\
\midrule
\endfirsthead
\toprule
Task & Prev. & EveryQuery & Autoregressive \\
\midrule
\endhead
\midrule \multicolumn{4}{r}{\emph{continued on next page}} \\
\endfoot
\bottomrule
\endlastfoot
Visit\bs{}OP within 365d \textbar{} no TIMELINE\bs{}\bs{}END within 365d, no MEDS\_DEATH within 365d & 0.730 & 0.543 & \textbf{0.720} \\
Visit\bs{}ER within 365d \textbar{} no TIMELINE\bs{}\bs{}END within 365d, no MEDS\_DEATH within 365d & 0.114 & 0.728 & \textbf{0.821} \\
MEDS\_DEATH within 1825d & 0.028 & \textbf{0.887} & 0.838 \\
Visit\bs{}ER within 30d \textbar{} no TIMELINE\bs{}\bs{}END within 30d, no MEDS\_DEATH within 30d & 0.025 & 0.662 & \textbf{0.762} \\
MEDS\_DEATH within 365d & 0.010 & \textbf{0.921} & 0.855 \\
\end{longtable}

\begin{longtable}{>{\raggedright\arraybackslash\ttfamily}p{0.62\linewidth} r r r}
\caption{AMC outpatient visit, heart-failure codes ($n=36$ tasks).}\label{tab:app-amc-hf}\\
\toprule
Task & Prev. & \multicolumn{2}{c}{AUROC} \\
\cmidrule(lr){3-4}
 & & EveryQuery & Autoregressive \\
\midrule
\endfirsthead
\toprule
Task & Prev. & EveryQuery & Autoregressive \\
\midrule
\endhead
\midrule \multicolumn{4}{r}{\emph{continued on next page}} \\
\endfoot
\bottomrule
\endlastfoot
LOINC\bs{}2823-3\bs{}\bs{}ANY within 90d \textbar{} no TIMELINE\bs{}\bs{}END within 90d, no MEDS\_DEATH within 90d & 0.623 & 0.799 & \textbf{0.827} \\
LOINC\bs{}2823-3\bs{}\bs{}UCUM\bs{}mmol\bs{}L\bs{}\bs{}value\_[5.0,inf) within 365d \textbar{} no TIMELINE\bs{}\bs{}END within 365d, no MEDS\_DEATH within 365d & 0.354 & \textbf{0.769} & 0.757 \\
ICD10CM\bs{}I48.91 within 1095d \textbar{} no TIMELINE\bs{}\bs{}END within 1095d & 0.284 & \textbf{0.828} & 0.799 \\
Visit\bs{}IP within 365d \textbar{} no TIMELINE\bs{}\bs{}END within 365d, no MEDS\_DEATH within 365d, no RxNorm\bs{}1373473 within 30d & 0.222 & \textbf{0.741} & 0.697 \\
Visit\bs{}IP within 365d \textbar{} no TIMELINE\bs{}\bs{}END within 365d, no MEDS\_DEATH within 365d, no RxNorm\bs{}1545658 within 30d & 0.222 & \textbf{0.741} & 0.697 \\
Visit\bs{}IP within 365d \textbar{} no TIMELINE\bs{}\bs{}END within 365d, no MEDS\_DEATH within 365d, no RxNorm\bs{}1545664 within 30d & 0.222 & \textbf{0.741} & 0.697 \\
Visit\bs{}IP within 365d \textbar{} no TIMELINE\bs{}\bs{}END within 365d, no MEDS\_DEATH within 365d, no RxNorm\bs{}1545666 within 30d & 0.222 & \textbf{0.741} & 0.697 \\
Visit\bs{}IP within 365d \textbar{} no TIMELINE\bs{}\bs{}END within 365d, no MEDS\_DEATH within 365d, no RxNorm\bs{}1373469 within 30d & 0.222 & \textbf{0.741} & 0.697 \\
Visit\bs{}IP within 365d \textbar{} no TIMELINE\bs{}\bs{}END within 365d, no MEDS\_DEATH within 365d, no RxNorm\bs{}1486977 within 30d & 0.222 & \textbf{0.741} & 0.697 \\
Visit\bs{}IP within 365d \textbar{} no TIMELINE\bs{}\bs{}END within 365d, no MEDS\_DEATH within 365d, no RxNorm\bs{}1486981 within 30d & 0.222 & \textbf{0.741} & 0.697 \\
Visit\bs{}IP within 365d \textbar{} no TIMELINE\bs{}\bs{}END within 365d, no MEDS\_DEATH within 365d, no RxNorm\bs{}1488569 within 30d & 0.222 & \textbf{0.741} & 0.697 \\
Visit\bs{}IP within 365d \textbar{} no TIMELINE\bs{}\bs{}END within 365d, no MEDS\_DEATH within 365d, no RxNorm\bs{}1545668 within 30d & 0.222 & \textbf{0.741} & 0.697 \\
Visit\bs{}IP within 365d \textbar{} no TIMELINE\bs{}\bs{}END within 365d, no MEDS\_DEATH within 365d & 0.222 & \textbf{0.740} & 0.697 \\
MEDS\_DEATH within 1825d & 0.168 & \textbf{0.773} & 0.711 \\
LOINC\bs{}2160-0\bs{}\bs{}UCUM\bs{}mg\bs{}dL\bs{}\bs{}value\_[2.34,inf) within 365d \textbar{} no TIMELINE\bs{}\bs{}END within 365d, no MEDS\_DEATH within 365d & 0.141 & \textbf{0.924} & 0.883 \\
Visit\bs{}ER within 90d \textbar{} no TIMELINE\bs{}\bs{}END within 90d, no MEDS\_DEATH within 90d & 0.119 & \textbf{0.728} & 0.716 \\
Visit\bs{}IP within 90d \textbar{} no TIMELINE\bs{}\bs{}END within 90d, no MEDS\_DEATH within 90d & 0.111 & \textbf{0.758} & 0.736 \\
LOINC\bs{}33762-6\bs{}\bs{}ANY within 365d \textbar{} no TIMELINE\bs{}\bs{}END within 365d, no MEDS\_DEATH within 365d & 0.109 & \textbf{0.909} & 0.803 \\
ICD10CM\bs{}N18.30 within 1095d \textbar{} no TIMELINE\bs{}\bs{}END within 1095d & 0.109 & \textbf{0.869} & 0.700 \\
ICD10CM\bs{}N18.3 within 1095d \textbar{} no TIMELINE\bs{}\bs{}END within 1095d & 0.083 & \textbf{0.912} & 0.733 \\
RxNorm\bs{}313096 within 365d \textbar{} no TIMELINE\bs{}\bs{}END within 365d, no MEDS\_DEATH within 365d & 0.068 & \textbf{0.819} & 0.789 \\
ICD10CM\bs{}I48.0 within 1095d \textbar{} no TIMELINE\bs{}\bs{}END within 1095d & 0.068 & \textbf{0.915} & 0.765 \\
MEDS\_DEATH within 365d & 0.049 & \textbf{0.817} & 0.744 \\
ICD10CM\bs{}I48.2 within 1095d \textbar{} no TIMELINE\bs{}\bs{}END within 1095d & 0.021 & \textbf{0.913} & 0.721 \\
RxNorm\bs{}200820 within 365d \textbar{} no TIMELINE\bs{}\bs{}END within 365d, no MEDS\_DEATH within 365d & 0.016 & \textbf{0.852} & 0.695 \\
ICD10CM\bs{}I48.20 within 1095d \textbar{} no TIMELINE\bs{}\bs{}END within 1095d & 0.014 & \textbf{0.935} & 0.737 \\
RxNorm\bs{}351256 within 365d \textbar{} no TIMELINE\bs{}\bs{}END within 365d, no MEDS\_DEATH within 365d & 0.013 & \textbf{0.904} & 0.751 \\
ICD10CM\bs{}I48.1 within 1095d \textbar{} no TIMELINE\bs{}\bs{}END within 1095d & 0.011 & \textbf{0.906} & 0.649 \\
ICD10CM\bs{}I48.19 within 1095d \textbar{} no TIMELINE\bs{}\bs{}END within 1095d & 0.010 & \textbf{0.939} & 0.661 \\
ICD10CM\bs{}N18.32 within 1095d \textbar{} no TIMELINE\bs{}\bs{}END within 1095d & 0.010 & \textbf{0.965} & 0.617 \\
ICD10PCS\bs{}0JH606Z within 730d \textbar{} no TIMELINE\bs{}\bs{}END within 730d & 0.009 & \textbf{0.726} & 0.540 \\
ICD10CM\bs{}N18.31 within 1095d \textbar{} no TIMELINE\bs{}\bs{}END within 1095d & 0.009 & \textbf{0.955} & 0.629 \\
RxNorm\bs{}198223 within 365d \textbar{} no TIMELINE\bs{}\bs{}END within 365d, no MEDS\_DEATH within 365d & 0.009 & \textbf{0.825} & 0.680 \\
ICD10PCS\bs{}02HK3JZ within 730d \textbar{} no TIMELINE\bs{}\bs{}END within 730d & 0.008 & \textbf{0.686} & 0.531 \\
ICD10CM\bs{}I48.11 within 1095d \textbar{} no TIMELINE\bs{}\bs{}END within 1095d & 0.007 & \textbf{0.961} & 0.675 \\
ICD10CM\bs{}I48.21 within 1095d \textbar{} no TIMELINE\bs{}\bs{}END within 1095d & 0.006 & \textbf{0.951} & 0.672 \\
\end{longtable}

\begin{longtable}{>{\raggedright\arraybackslash\ttfamily}p{0.62\linewidth} r r r}
\caption{AMC outpatient visit, type 2 diabetes codes ($n=59$ tasks).}\label{tab:app-amc-t2d}\\
\toprule
Task & Prev. & \multicolumn{2}{c}{AUROC} \\
\cmidrule(lr){3-4}
 & & EveryQuery & Autoregressive \\
\midrule
\endfirsthead
\toprule
Task & Prev. & EveryQuery & Autoregressive \\
\midrule
\endhead
\midrule \multicolumn{4}{r}{\emph{continued on next page}} \\
\endfoot
\bottomrule
\endlastfoot
LOINC\bs{}4548-4\bs{}\bs{}ANY within 180d \textbar{} no TIMELINE\bs{}\bs{}END within 180d, no MEDS\_DEATH within 180d & 0.301 & 0.866 & \textbf{0.867} \\
ICD10CM\bs{}I50.814 within 1095d \textbar{} no TIMELINE\bs{}\bs{}END within 1095d & 0.151 & \textbf{0.898} & 0.821 \\
LOINC\bs{}2160-0\bs{}\bs{}UCUM\bs{}mg\bs{}dL\bs{}\bs{}value\_[2.34,inf) within 730d \textbar{} no TIMELINE\bs{}\bs{}END within 730d & 0.137 & \textbf{0.930} & 0.841 \\
MEDS\_DEATH within 1825d & 0.121 & \textbf{0.813} & 0.734 \\
ICD10CM\bs{}E11.319 within 1095d \textbar{} no TIMELINE\bs{}\bs{}END within 1095d & 0.116 & \textbf{0.832} & 0.764 \\
LOINC\bs{}14959-1\bs{}\bs{}ANY within 365d \textbar{} no TIMELINE\bs{}\bs{}END within 365d, no MEDS\_DEATH within 365d & 0.107 & \textbf{0.880} & 0.837 \\
LOINC\bs{}4548-4\bs{}\bs{}UCUM\bs{}\%\bs{}\bs{}value\_[8.9,inf) within 365d \textbar{} no TIMELINE\bs{}\bs{}END within 365d, no MEDS\_DEATH within 365d & 0.094 & \textbf{0.870} & 0.864 \\
ICD10CM\bs{}N18.30 within 1095d \textbar{} no TIMELINE\bs{}\bs{}END within 1095d & 0.090 & \textbf{0.895} & 0.721 \\
ICD10CM\bs{}N18.3 within 1095d \textbar{} no TIMELINE\bs{}\bs{}END within 1095d & 0.073 & \textbf{0.930} & 0.763 \\
ICD10CM\bs{}I50.9 within 1095d \textbar{} no TIMELINE\bs{}\bs{}END within 1095d & 0.071 & \textbf{0.925} & 0.774 \\
ICD10CM\bs{}I50.22 within 1095d \textbar{} no TIMELINE\bs{}\bs{}END within 1095d & 0.070 & \textbf{0.881} & 0.742 \\
RxNorm\bs{}617310 within 365d \textbar{} no TIMELINE\bs{}\bs{}END within 365d, no MEDS\_DEATH within 365d & 0.064 & \textbf{0.709} & 0.694 \\
RxNorm\bs{}617320 within 365d \textbar{} no TIMELINE\bs{}\bs{}END within 365d, no MEDS\_DEATH within 365d & 0.050 & \textbf{0.800} & 0.755 \\
RxNorm\bs{}617311 within 365d \textbar{} no TIMELINE\bs{}\bs{}END within 365d, no MEDS\_DEATH within 365d & 0.050 & \textbf{0.794} & 0.731 \\
RxNorm\bs{}312961 within 365d \textbar{} no TIMELINE\bs{}\bs{}END within 365d, no MEDS\_DEATH within 365d & 0.044 & \textbf{0.761} & 0.755 \\
ICD10CM\bs{}I50.20 within 1095d \textbar{} no TIMELINE\bs{}\bs{}END within 1095d & 0.043 & \textbf{0.896} & 0.693 \\
RxNorm\bs{}259255 within 365d \textbar{} no TIMELINE\bs{}\bs{}END within 365d, no MEDS\_DEATH within 365d & 0.042 & \textbf{0.847} & 0.742 \\
ICD10CM\bs{}I50.32 within 1095d \textbar{} no TIMELINE\bs{}\bs{}END within 1095d & 0.041 & \textbf{0.853} & 0.656 \\
RxNorm\bs{}198211 within 365d \textbar{} no TIMELINE\bs{}\bs{}END within 365d, no MEDS\_DEATH within 365d & 0.039 & \textbf{0.797} & 0.778 \\
ICD10CM\bs{}I50.23 within 1095d \textbar{} no TIMELINE\bs{}\bs{}END within 1095d & 0.034 & \textbf{0.884} & 0.636 \\
ICD10CM\bs{}I50.30 within 1095d \textbar{} no TIMELINE\bs{}\bs{}END within 1095d & 0.034 & \textbf{0.842} & 0.618 \\
RxNorm\bs{}617314 within 365d \textbar{} no TIMELINE\bs{}\bs{}END within 365d, no MEDS\_DEATH within 365d & 0.031 & \textbf{0.716} & 0.704 \\
RxNorm\bs{}262095 within 365d \textbar{} no TIMELINE\bs{}\bs{}END within 365d, no MEDS\_DEATH within 365d & 0.026 & \textbf{0.842} & 0.772 \\
RxNorm\bs{}617318 within 365d \textbar{} no TIMELINE\bs{}\bs{}END within 365d, no MEDS\_DEATH within 365d & 0.026 & \textbf{0.743} & 0.722 \\
RxNorm\bs{}617312 within 365d \textbar{} no TIMELINE\bs{}\bs{}END within 365d, no MEDS\_DEATH within 365d & 0.026 & 0.711 & \textbf{0.744} \\
ICD10CM\bs{}I50.810 within 1095d \textbar{} no TIMELINE\bs{}\bs{}END within 1095d & 0.024 & \textbf{0.869} & 0.636 \\
ICD10CM\bs{}E11.3593 within 1095d \textbar{} no TIMELINE\bs{}\bs{}END within 1095d & 0.021 & \textbf{0.907} & 0.735 \\
ICD10CM\bs{}I50.33 within 1095d \textbar{} no TIMELINE\bs{}\bs{}END within 1095d & 0.020 & \textbf{0.886} & 0.564 \\
RxNorm\bs{}904475 within 365d \textbar{} no TIMELINE\bs{}\bs{}END within 365d, no MEDS\_DEATH within 365d & 0.020 & \textbf{0.812} & 0.626 \\
ICD10CM\bs{}I50.21 within 1095d \textbar{} no TIMELINE\bs{}\bs{}END within 1095d & 0.019 & \textbf{0.837} & 0.575 \\
RxNorm\bs{}904467 within 365d \textbar{} no TIMELINE\bs{}\bs{}END within 365d, no MEDS\_DEATH within 365d & 0.016 & \textbf{0.863} & 0.618 \\
RxNorm\bs{}314231 within 365d \textbar{} no TIMELINE\bs{}\bs{}END within 365d, no MEDS\_DEATH within 365d & 0.015 & \textbf{0.732} & 0.690 \\
RxNorm\bs{}859753 within 365d \textbar{} no TIMELINE\bs{}\bs{}END within 365d, no MEDS\_DEATH within 365d & 0.014 & \textbf{0.768} & 0.724 \\
RxNorm\bs{}859749 within 365d \textbar{} no TIMELINE\bs{}\bs{}END within 365d, no MEDS\_DEATH within 365d & 0.014 & \textbf{0.710} & 0.682 \\
CPT4\bs{}92250 within 365d \textbar{} no TIMELINE\bs{}\bs{}END within 365d, no MEDS\_DEATH within 365d & 0.014 & \textbf{0.773} & 0.634 \\
ICD10CM\bs{}I50.31 within 1095d \textbar{} no TIMELINE\bs{}\bs{}END within 1095d & 0.013 & \textbf{0.854} & 0.572 \\
ICD10CM\bs{}N18.31 within 1095d \textbar{} no TIMELINE\bs{}\bs{}END within 1095d & 0.013 & \textbf{0.959} & 0.653 \\
RxNorm\bs{}104491 within 365d \textbar{} no TIMELINE\bs{}\bs{}END within 365d, no MEDS\_DEATH within 365d & 0.011 & \textbf{0.835} & 0.719 \\
RxNorm\bs{}859421 within 365d \textbar{} no TIMELINE\bs{}\bs{}END within 365d, no MEDS\_DEATH within 365d & 0.011 & \textbf{0.874} & 0.800 \\
RxNorm\bs{}152923 within 365d \textbar{} no TIMELINE\bs{}\bs{}END within 365d, no MEDS\_DEATH within 365d & 0.010 & \textbf{0.853} & 0.793 \\
ICD10CM\bs{}I50.43 within 1095d \textbar{} no TIMELINE\bs{}\bs{}END within 1095d & 0.010 & \textbf{0.896} & 0.545 \\
ICD10CM\bs{}N18.32 within 1095d \textbar{} no TIMELINE\bs{}\bs{}END within 1095d & 0.010 & \textbf{0.965} & 0.688 \\
ICD10CM\bs{}I50.1 within 1095d \textbar{} no TIMELINE\bs{}\bs{}END within 1095d & 0.009 & \textbf{0.930} & 0.600 \\
RxNorm\bs{}859747 within 365d \textbar{} no TIMELINE\bs{}\bs{}END within 365d, no MEDS\_DEATH within 365d & 0.009 & \textbf{0.866} & 0.695 \\
ICD10CM\bs{}I50.42 within 1095d \textbar{} no TIMELINE\bs{}\bs{}END within 1095d & 0.009 & \textbf{0.873} & 0.602 \\
RxNorm\bs{}859751 within 365d \textbar{} no TIMELINE\bs{}\bs{}END within 365d, no MEDS\_DEATH within 365d & 0.009 & \textbf{0.823} & 0.705 \\
RxNorm\bs{}897126 within 365d \textbar{} no TIMELINE\bs{}\bs{}END within 365d, no MEDS\_DEATH within 365d & 0.009 & \textbf{0.860} & 0.735 \\
RxNorm\bs{}859419 within 365d \textbar{} no TIMELINE\bs{}\bs{}END within 365d, no MEDS\_DEATH within 365d & 0.007 & \textbf{0.860} & 0.816 \\
RxNorm\bs{}200345 within 365d \textbar{} no TIMELINE\bs{}\bs{}END within 365d, no MEDS\_DEATH within 365d & 0.007 & \textbf{0.832} & 0.768 \\
RxNorm\bs{}1545664 within 365d \textbar{} no TIMELINE\bs{}\bs{}END within 365d, no MEDS\_DEATH within 365d & 0.007 & \textbf{0.932} & 0.770 \\
RxNorm\bs{}859426 within 365d \textbar{} no TIMELINE\bs{}\bs{}END within 365d, no MEDS\_DEATH within 365d & 0.007 & \textbf{0.758} & 0.697 \\
ICD10CM\bs{}E11.621 within 730d \textbar{} no TIMELINE\bs{}\bs{}END within 730d & 0.007 & \textbf{0.928} & 0.656 \\
RxNorm\bs{}904458 within 365d \textbar{} no TIMELINE\bs{}\bs{}END within 365d, no MEDS\_DEATH within 365d & 0.006 & \textbf{0.796} & 0.661 \\
ICD10CM\bs{}E11.649 within 365d \textbar{} no TIMELINE\bs{}\bs{}END within 365d, no MEDS\_DEATH within 365d & 0.006 & \textbf{0.926} & 0.547 \\
ICD10CM\bs{}E11.3293 within 1095d \textbar{} no TIMELINE\bs{}\bs{}END within 1095d & 0.006 & \textbf{0.859} & 0.579 \\
RxNorm\bs{}859424 within 365d \textbar{} no TIMELINE\bs{}\bs{}END within 365d, no MEDS\_DEATH within 365d & 0.006 & \textbf{0.921} & 0.660 \\
RxNorm\bs{}904477 within 365d \textbar{} no TIMELINE\bs{}\bs{}END within 365d, no MEDS\_DEATH within 365d & 0.006 & \textbf{0.818} & 0.679 \\
RxNorm\bs{}904469 within 365d \textbar{} no TIMELINE\bs{}\bs{}END within 365d, no MEDS\_DEATH within 365d & 0.005 & \textbf{0.875} & 0.546 \\
RxNorm\bs{}1551300 within 365d \textbar{} no TIMELINE\bs{}\bs{}END within 365d, no MEDS\_DEATH within 365d & 0.004 & \textbf{0.912} & 0.702 \\
\end{longtable}

\begin{longtable}{>{\raggedright\arraybackslash\ttfamily}p{0.62\linewidth} r r r}
\caption{MIMIC-IV ICU, 24h after admission ($n=38$ tasks).}\label{tab:app-mimic}\\
\toprule
Task & Prev. & \multicolumn{2}{c}{AUROC} \\
\cmidrule(lr){3-4}
 & & EveryQuery & Autoregressive \\
\midrule
\endfirsthead
\toprule
Task & Prev. & EveryQuery & Autoregressive \\
\midrule
\endhead
\midrule \multicolumn{4}{r}{\emph{continued on next page}} \\
\endfoot
\bottomrule
\endlastfoot
ICU\_DISCHARGE within 7d \textbar{} no TIMELINE\bs{}\bs{}END within 7d, no MEDS\_DEATH before ICU\_DISCHARGE & 0.882 & \textbf{0.782} & 0.710 \\
ICU\_DISCHARGE within 3d \textbar{} no TIMELINE\bs{}\bs{}END within 3d, no MEDS\_DEATH before ICU\_DISCHARGE & 0.712 & \textbf{0.796} & 0.629 \\
ICU\_DISCHARGE within 2d \textbar{} no TIMELINE\bs{}\bs{}END within 2d, no MEDS\_DEATH before ICU\_DISCHARGE & 0.599 & \textbf{0.798} & 0.595 \\
ICU\_DISCHARGE within 1d \textbar{} no TIMELINE\bs{}\bs{}END within 1d, no MEDS\_DEATH before ICU\_DISCHARGE & 0.390 & \textbf{0.798} & 0.564 \\
MEDS\_DEATH within 365d & 0.293 & \textbf{0.831} & 0.727 \\
MEDS\_DEATH within 90d & 0.207 & \textbf{0.849} & 0.747 \\
LAB\bs{}\bs{}RESULT\bs{}\bs{}50983\bs{}\bs{}mEq\bs{}L\bs{}\bs{}value\_[-inf,133.0) within 2d \textbar{} no TIMELINE\bs{}\bs{}END within 2d, no MEDS\_DEATH within 2d & 0.150 & \textbf{0.743} & 0.668 \\
MEDS\_DEATH within 28d & 0.146 & \textbf{0.874} & 0.771 \\
LAB\bs{}\bs{}RESULT\bs{}\bs{}50912\bs{}\bs{}mg\bs{}dL\bs{}\bs{}value\_[2.3,inf) within 3d \textbar{} no TIMELINE\bs{}\bs{}END within 3d, no MEDS\_DEATH within 3d & 0.143 & \textbf{0.899} & 0.789 \\
MEDICATION\bs{}\bs{}START\bs{}\bs{}Vancomycin within 1d \textbar{} no TIMELINE\bs{}\bs{}END within 1d, no MEDS\_DEATH within 1d & 0.135 & \textbf{0.749} & 0.662 \\
LAB\bs{}\bs{}RESULT\bs{}\bs{}50912\bs{}\bs{}mg\bs{}dL\bs{}\bs{}value\_[2.3,inf) within 2d \textbar{} no TIMELINE\bs{}\bs{}END within 2d, no MEDS\_DEATH within 2d & 0.134 & \textbf{0.903} & 0.797 \\
LAB\bs{}\bs{}RESULT\bs{}\bs{}50882\bs{}\bs{}mEq\bs{}L\bs{}\bs{}value\_[-inf,20.0) within 1d \textbar{} no TIMELINE\bs{}\bs{}END within 1d, no MEDS\_DEATH within 1d & 0.134 & \textbf{0.816} & 0.755 \\
LAB\bs{}\bs{}RESULT\bs{}\bs{}51222\bs{}\bs{}g\bs{}dL\bs{}\bs{}value\_[-inf,7.9) within 1d \textbar{} no TIMELINE\bs{}\bs{}END within 1d, no MEDS\_DEATH within 1d & 0.132 & \textbf{0.801} & 0.727 \\
LAB\bs{}\bs{}RESULT\bs{}\bs{}50971\bs{}\bs{}mEq\bs{}L\bs{}\bs{}value\_[-inf,3.5) within 1d \textbar{} no TIMELINE\bs{}\bs{}END within 1d, no MEDS\_DEATH within 1d & 0.125 & \textbf{0.711} & 0.658 \\
LAB\bs{}\bs{}RESULT\bs{}\bs{}51265\bs{}\bs{}K\bs{}uL\bs{}\bs{}value\_[-inf,86.0) within 2d \textbar{} no TIMELINE\bs{}\bs{}END within 2d, no MEDS\_DEATH within 2d & 0.123 & \textbf{0.849} & 0.737 \\
LAB\bs{}\bs{}RESULT\bs{}\bs{}50931\bs{}\bs{}mg\bs{}dL\bs{}\bs{}value\_[191.0,inf) within 1d \textbar{} no TIMELINE\bs{}\bs{}END within 1d, no MEDS\_DEATH within 1d & 0.116 & \textbf{0.823} & 0.732 \\
MEDS\_DEATH before ICU\_DISCHARGE & 0.072 & \textbf{0.860} & 0.810 \\
LAB\bs{}\bs{}RESULT\bs{}\bs{}50885\bs{}\bs{}mg\bs{}dL\bs{}\bs{}value\_[2.6,inf) within 3d \textbar{} no TIMELINE\bs{}\bs{}END within 3d, no MEDS\_DEATH within 3d & 0.072 & \textbf{0.904} & 0.739 \\
LAB\bs{}\bs{}RESULT\bs{}\bs{}50821\bs{}\bs{}mm Hg\bs{}\bs{}value\_[-inf,43.0) within 1d \textbar{} no TIMELINE\bs{}\bs{}END within 1d, no MEDS\_DEATH within 1d & 0.056 & \textbf{0.859} & 0.711 \\
LAB\bs{}\bs{}RESULT\bs{}\bs{}51237\bs{}\bs{}UNK\bs{}\bs{}value\_[2.6,inf) within 2d \textbar{} no TIMELINE\bs{}\bs{}END within 2d, no MEDS\_DEATH within 2d & 0.053 & \textbf{0.790} & 0.644 \\
MEDICATION\bs{}\bs{}START\bs{}\bs{}CefePIME within 1d \textbar{} no TIMELINE\bs{}\bs{}END within 1d, no MEDS\_DEATH within 1d & 0.043 & \textbf{0.807} & 0.650 \\
MEDS\_DEATH within 3d & 0.040 & \textbf{0.939} & 0.833 \\
LAB\bs{}\bs{}RESULT\bs{}\bs{}50820\bs{}\bs{}units\bs{}\bs{}value\_[-inf,7.26) within 0.5d \textbar{} no TIMELINE\bs{}\bs{}END within 0.5d, no MEDS\_DEATH within 0.5d & 0.024 & \textbf{0.904} & 0.831 \\
MEDICATION\bs{}\bs{}START\bs{}\bs{}Piperacillin-Tazobactam within 1d \textbar{} no TIMELINE\bs{}\bs{}END within 1d, no MEDS\_DEATH within 1d & 0.024 & \textbf{0.833} & 0.648 \\
PROCEDURE\bs{}\bs{}START\bs{}\bs{}225792 within 1d \textbar{} no TIMELINE\bs{}\bs{}END within 1d, no MEDS\_DEATH within 1d & 0.019 & \textbf{0.785} & 0.499 \\
LAB\bs{}\bs{}RESULT\bs{}\bs{}50813\bs{}\bs{}mmol\bs{}L\bs{}\bs{}value\_[4.0,inf) within 0.5d \textbar{} no TIMELINE\bs{}\bs{}END within 0.5d, no MEDS\_DEATH within 0.5d & 0.019 & \textbf{0.923} & 0.898 \\
MEDS\_DEATH within 1d & 0.016 & \textbf{0.959} & 0.894 \\
LAB\bs{}\bs{}RESULT\bs{}\bs{}50813\bs{}\bs{}mmol\bs{}L\bs{}\bs{}value\_[4.0,inf) within 0.25d \textbar{} no TIMELINE\bs{}\bs{}END within 0.25d, no MEDS\_DEATH within 0.25d & 0.015 & \textbf{0.945} & 0.938 \\
INFUSION\_START\bs{}\bs{}221906\bs{}\bs{}value\_[0.32051384,inf) within 1d \textbar{} no TIMELINE\bs{}\bs{}END within 1d, no MEDS\_DEATH within 1d & 0.015 & \textbf{0.952} & 0.763 \\
INFUSION\_START\bs{}\bs{}221906\bs{}\bs{}value\_[0.32051384,inf) within 0.5d \textbar{} no TIMELINE\bs{}\bs{}END within 0.5d, no MEDS\_DEATH within 0.5d & 0.013 & \textbf{0.958} & 0.924 \\
MEDICATION\bs{}\bs{}START\bs{}\bs{}Levofloxacin within 1d \textbar{} no TIMELINE\bs{}\bs{}END within 1d, no MEDS\_DEATH within 1d & 0.012 & \textbf{0.840} & 0.604 \\
PROCEDURE\bs{}\bs{}START\bs{}\bs{}225441 within 2d \textbar{} no TIMELINE\bs{}\bs{}END within 2d, no MEDS\_DEATH within 2d & 0.011 & \textbf{0.931} & 0.600 \\
PROCEDURE\bs{}\bs{}START\bs{}\bs{}224385 within 1d \textbar{} no TIMELINE\bs{}\bs{}END within 1d, no MEDS\_DEATH within 1d & 0.011 & \textbf{0.813} & 0.498 \\
PROCEDURE\bs{}\bs{}START\bs{}\bs{}225792 within 0.5d \textbar{} no TIMELINE\bs{}\bs{}END within 0.5d, no MEDS\_DEATH within 0.5d & 0.011 & \textbf{0.814} & 0.638 \\
MEDICATION\bs{}\bs{}START\bs{}\bs{}Ciprofloxacin HCl within 1d \textbar{} no TIMELINE\bs{}\bs{}END within 1d, no MEDS\_DEATH within 1d & 0.010 & \textbf{0.776} & 0.522 \\
MEDICATION\bs{}\bs{}START\bs{}\bs{}Meropenem within 1d \textbar{} no TIMELINE\bs{}\bs{}END within 1d, no MEDS\_DEATH within 1d & 0.008 & \textbf{0.897} & 0.590 \\
INFUSION\_START\bs{}\bs{}221749\bs{}\bs{}value\_[3.0013728,inf) within 1d \textbar{} no TIMELINE\bs{}\bs{}END within 1d, no MEDS\_DEATH within 1d & 0.007 & \textbf{0.919} & 0.755 \\
PROCEDURE\bs{}\bs{}START\bs{}\bs{}225794 within 1d \textbar{} no TIMELINE\bs{}\bs{}END within 1d, no MEDS\_DEATH within 1d & 0.006 & \textbf{0.829} & 0.555 \\
\end{longtable}

\begin{longtable}{>{\raggedright\arraybackslash\ttfamily}p{0.62\linewidth} r r r}
\caption{NWICU ICU, 24h after admission ($n=27$ tasks).}\label{tab:app-nwicu}\\
\toprule
Task & Prev. & \multicolumn{2}{c}{AUROC} \\
\cmidrule(lr){3-4}
 & & EveryQuery & Autoregressive \\
\midrule
\endfirsthead
\toprule
Task & Prev. & EveryQuery & Autoregressive \\
\midrule
\endhead
\midrule \multicolumn{4}{r}{\emph{continued on next page}} \\
\endfoot
\bottomrule
\endlastfoot
ICU\_DISCHARGE within 7d \textbar{} no TIMELINE\bs{}\bs{}END within 7d, no MEDS\_DEATH before ICU\_DISCHARGE & 0.875 & \textbf{0.721} & 0.528 \\
ICU\_DISCHARGE within 3d \textbar{} no TIMELINE\bs{}\bs{}END within 3d, no MEDS\_DEATH before ICU\_DISCHARGE & 0.686 & \textbf{0.706} & 0.619 \\
ICU\_DISCHARGE within 2d \textbar{} no TIMELINE\bs{}\bs{}END within 2d, no MEDS\_DEATH before ICU\_DISCHARGE & 0.565 & \textbf{0.688} & 0.661 \\
ICU\_DISCHARGE within 1d \textbar{} no TIMELINE\bs{}\bs{}END within 1d, no MEDS\_DEATH before ICU\_DISCHARGE & 0.369 & 0.676 & \textbf{0.693} \\
MEDS\_DEATH within 365d & 0.203 & \textbf{0.817} & 0.737 \\
MEDS\_DEATH within 90d & 0.162 & \textbf{0.850} & 0.781 \\
MEDS\_DEATH within 28d & 0.133 & \textbf{0.874} & 0.809 \\
PROCEDURE\bs{}\bs{}START\bs{}\bs{}787541 within 1d \textbar{} no TIMELINE\bs{}\bs{}END within 1d, no MEDS\_DEATH within 1d & 0.131 & 0.955 & \textbf{0.963} \\
PROCEDURE\bs{}\bs{}START\bs{}\bs{}792843 within 1d \textbar{} no TIMELINE\bs{}\bs{}END within 1d, no MEDS\_DEATH within 1d & 0.109 & 0.823 & \textbf{0.908} \\
LAB\bs{}\bs{}100001\bs{}\bs{}mg\bs{}dL\bs{}\bs{}value\_[236.0,inf) within 1d \textbar{} no TIMELINE\bs{}\bs{}END within 1d, no MEDS\_DEATH within 1d & 0.102 & \textbf{0.899} & 0.888 \\
LAB\bs{}\bs{}100002\bs{}\bs{}mg\bs{}dL\bs{}\bs{}value\_[3.27,inf) within 3d \textbar{} no TIMELINE\bs{}\bs{}END within 3d, no MEDS\_DEATH within 3d & 0.094 & \textbf{0.957} & 0.956 \\
PROCEDURE\bs{}\bs{}START\bs{}\bs{}787541 within 0.5d \textbar{} no TIMELINE\bs{}\bs{}END within 0.5d, no MEDS\_DEATH within 0.5d & 0.091 & 0.935 & \textbf{0.956} \\
LAB\bs{}\bs{}100002\bs{}\bs{}mg\bs{}dL\bs{}\bs{}value\_[3.27,inf) within 2d \textbar{} no TIMELINE\bs{}\bs{}END within 2d, no MEDS\_DEATH within 2d & 0.089 & \textbf{0.963} & 0.962 \\
PROCEDURE\bs{}\bs{}START\bs{}\bs{}792843 within 0.5d \textbar{} no TIMELINE\bs{}\bs{}END within 0.5d, no MEDS\_DEATH within 0.5d & 0.075 & 0.787 & \textbf{0.898} \\
LAB\bs{}\bs{}100007\bs{}\bs{}g\bs{}dL\bs{}\bs{}value\_[-inf,7.4) within 1d \textbar{} no TIMELINE\bs{}\bs{}END within 1d, no MEDS\_DEATH within 1d & 0.070 & \textbf{0.920} & 0.918 \\
MEDS\_DEATH before HOSPITAL\_DISCHARGE & 0.070 & \textbf{0.839} & 0.783 \\
LAB\bs{}\bs{}100013\bs{}\bs{}mEq\bs{}L\bs{}\bs{}value\_[-inf,20.0) within 1d \textbar{} no TIMELINE\bs{}\bs{}END within 1d, no MEDS\_DEATH within 1d & 0.064 & 0.900 & \textbf{0.933} \\
LAB\bs{}\bs{}100011\bs{}\bs{}mEq\bs{}L\bs{}\bs{}value\_[-inf,3.4) within 1d \textbar{} no TIMELINE\bs{}\bs{}END within 1d, no MEDS\_DEATH within 1d & 0.062 & 0.841 & \textbf{0.860} \\
LAB\bs{}\bs{}100010\bs{}\bs{}mEq\bs{}L\bs{}\bs{}value\_[-inf,132.0) within 2d \textbar{} no TIMELINE\bs{}\bs{}END within 2d, no MEDS\_DEATH within 2d & 0.047 & 0.835 & \textbf{0.858} \\
LAB\bs{}\bs{}100020\bs{}\bs{}mg\bs{}dL\bs{}\bs{}value\_[3.0,inf) within 3d \textbar{} no TIMELINE\bs{}\bs{}END within 3d, no MEDS\_DEATH within 3d & 0.044 & \textbf{0.963} & 0.935 \\
MEDS\_DEATH before ICU\_DISCHARGE & 0.044 & \textbf{0.842} & 0.782 \\
LAB\bs{}\bs{}100011\bs{}\bs{}mmol\bs{}L\bs{}\bs{}value\_[-inf,3.4) within 1d \textbar{} no TIMELINE\bs{}\bs{}END within 1d, no MEDS\_DEATH within 1d & 0.042 & \textbf{0.893} & 0.870 \\
MEDS\_DEATH within 3d & 0.038 & \textbf{0.933} & 0.886 \\
LAB\bs{}\bs{}100013\bs{}\bs{}mmol\bs{}L\bs{}\bs{}value\_[-inf,20.0) within 1d \textbar{} no TIMELINE\bs{}\bs{}END within 1d, no MEDS\_DEATH within 1d & 0.032 & 0.926 & \textbf{0.932} \\
LAB\bs{}\bs{}100014\bs{}\bs{}10\^{}3\bs{}$\mu$L\bs{}\bs{}value\_[-inf,82.0) within 2d \textbar{} no TIMELINE\bs{}\bs{}END within 2d, no MEDS\_DEATH within 2d & 0.031 & \textbf{0.961} & 0.959 \\
LAB\bs{}\bs{}100010\bs{}\bs{}mmol\bs{}L\bs{}\bs{}value\_[-inf,132.0) within 2d \textbar{} no TIMELINE\bs{}\bs{}END within 2d, no MEDS\_DEATH within 2d & 0.029 & \textbf{0.915} & 0.912 \\
LAB\bs{}\bs{}100014\bs{}\bs{}K\bs{}UL\bs{}\bs{}value\_[-inf,56.0) within 2d \textbar{} no TIMELINE\bs{}\bs{}END within 2d, no MEDS\_DEATH within 2d & 0.026 & \textbf{0.969} & 0.957 \\
\end{longtable}

\clearpage
\section{Results on Time-to-Event Tasks}
\label{app:tte}

\subsection{Target codes and prediction horizons}
\label{app:tte_codes}
Table~\ref{tab:tte_codes} lists the target codes used for time-to-event evaluation in each dataset. For every code, we evaluate the survival curve at 16 horizons: 1, 2, 3, 5, 7, 10, 14, 21, 30, 45, 60, 90, 120, 180, 270 and 365 days.

\begin{table}[H]
\centering
\small
\caption{\textbf{Target codes for time-to-event evaluation.} Codes are written as they appear in each dataset's vocabulary.}
\label{tab:tte_codes}
\begin{tabular}{ll}
\toprule
\textbf{Dataset} & \textbf{Target code} \\
\midrule
MIMIC-IV & \texttt{MEDS\_DEATH} \\
 & \texttt{PROCEDURE//START//225792} \\
 & \texttt{ICU\_ADMISSION//Medical Intensive Care Unit (MICU)} \\
 & \texttt{MEDICATION//START//Norepinephrine} \\
 & \texttt{HOSPITAL\_DISCHARGE//HOME} \\
 & \texttt{INFUSION\_START//221289//value\_[0.35491893,inf)} \\
 & \texttt{INFUSION\_START//221749//value\_[3.0013728,inf)} \\
\midrule
NWICU & \texttt{DEATH} \\
 & \texttt{PROCEDURE//START//787541} \\
 & \texttt{PROCEDURE//START//792843} \\
 & \texttt{DIAGNOSIS//ICD//10//A419} \\
 & \texttt{HOSPITAL\_DISCHARGE} \\
\midrule
AMC & \texttt{MEDS\_DEATH} \\
 & \texttt{ICD10CM/I48.0} \\
 & \texttt{Visit/IP} \\
 & \texttt{Visit/ER} \\
 & \texttt{LOINC/2160-0//UCUM/mg/dL//value\_[2.34,inf)} \\
 & \texttt{LOINC/2823-3//UCUM/mmol/L//value\_[5.0,inf)} \\
 & \texttt{ICD10CM/N18.3} \\
 & \texttt{RxNorm/1373469} \\
 & \texttt{RxNorm/198222} \\
\bottomrule
\end{tabular}
\end{table}

\clearpage

\subsection{Survival curves}
\label{app:tte_curves}
\begin{figure}[H]
    \centering
    \includegraphics[width=\textwidth]{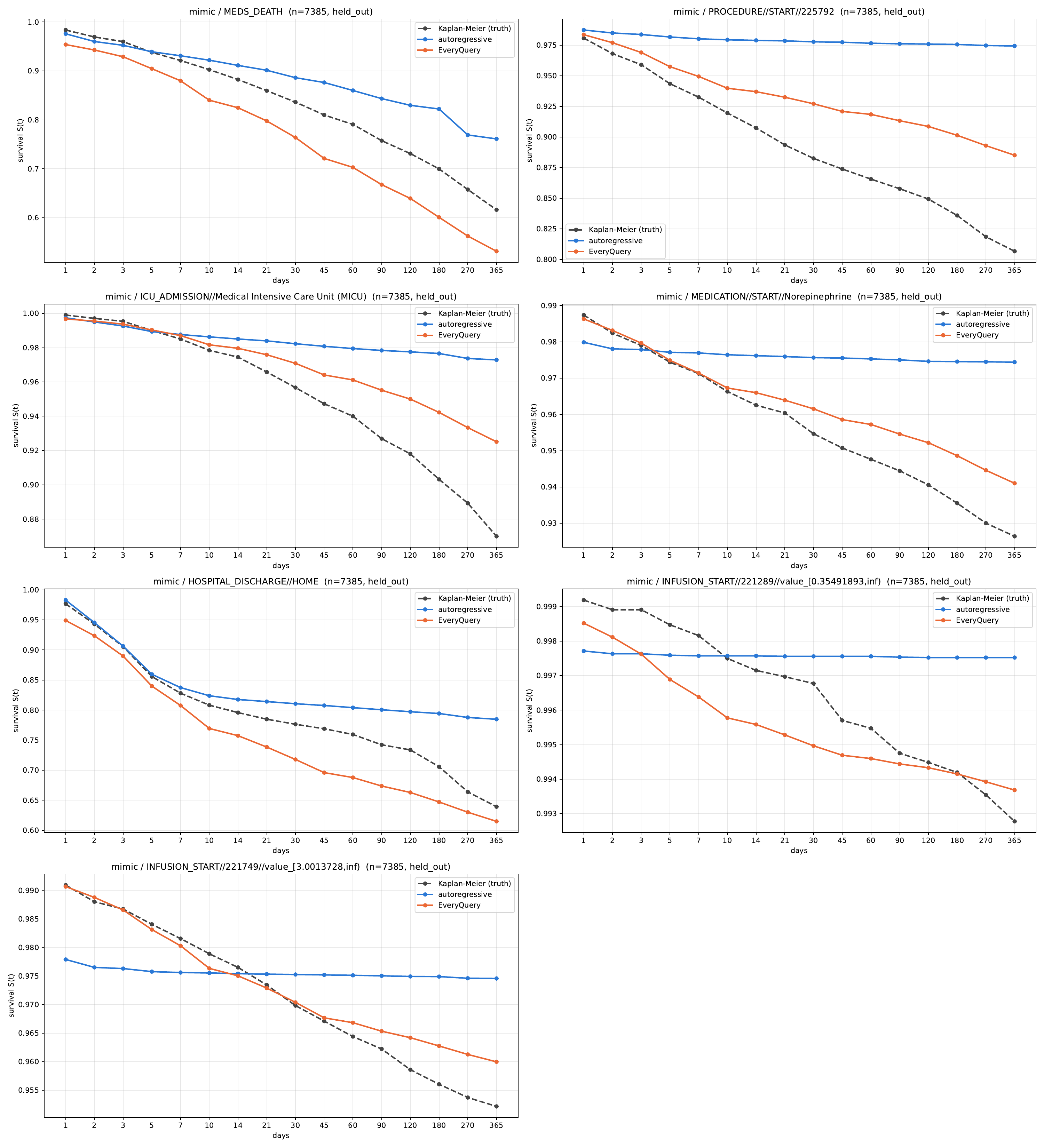}
\caption{\textbf{Survival curves on MIMIC-IV.} Each panel shows one target code (Table~\ref{tab:tte_codes}), comparing the Kaplan--Meier estimate on held-out patients (black, dashed) with the survival curves estimated by the autoregressive baseline (blue) and \model (orange) at each prediction horizon in days.}
    \label{fig:mimic-survival-curves}
\end{figure}

\begin{figure}[H]
    \centering
    \includegraphics[width=\textwidth]{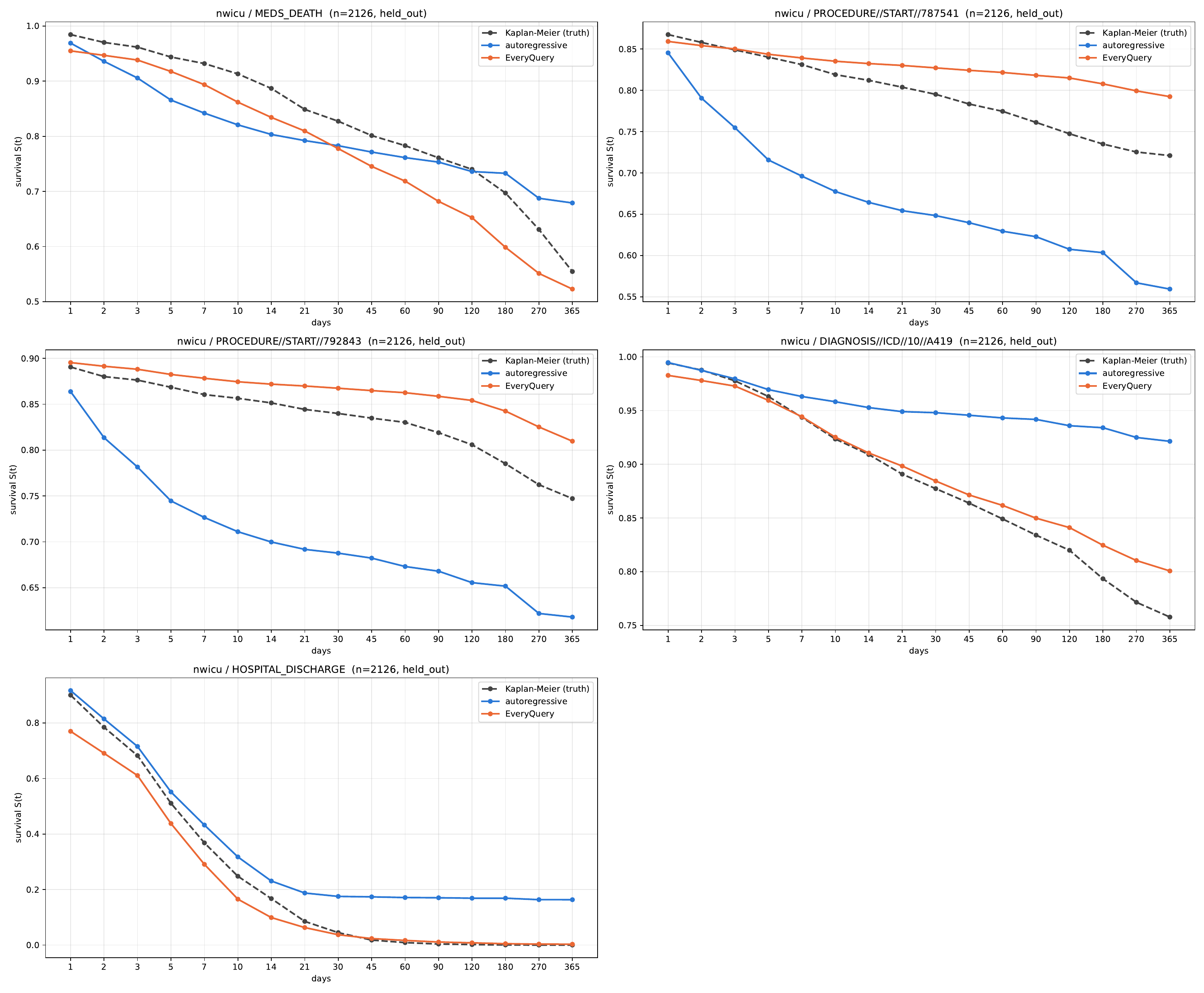}
\caption{\textbf{Survival curves on NWICU.} Each panel shows one target code (Table~\ref{tab:tte_codes}), comparing the Kaplan--Meier estimate on held-out patients (black, dashed) with the survival curves estimated by the autoregressive baseline (blue) and \model (orange) at each prediction horizon in days.}
    \label{fig:nwicu-survival-curves}
\end{figure}

\begin{figure}[H]
    \centering
    \includegraphics[width=\textwidth]{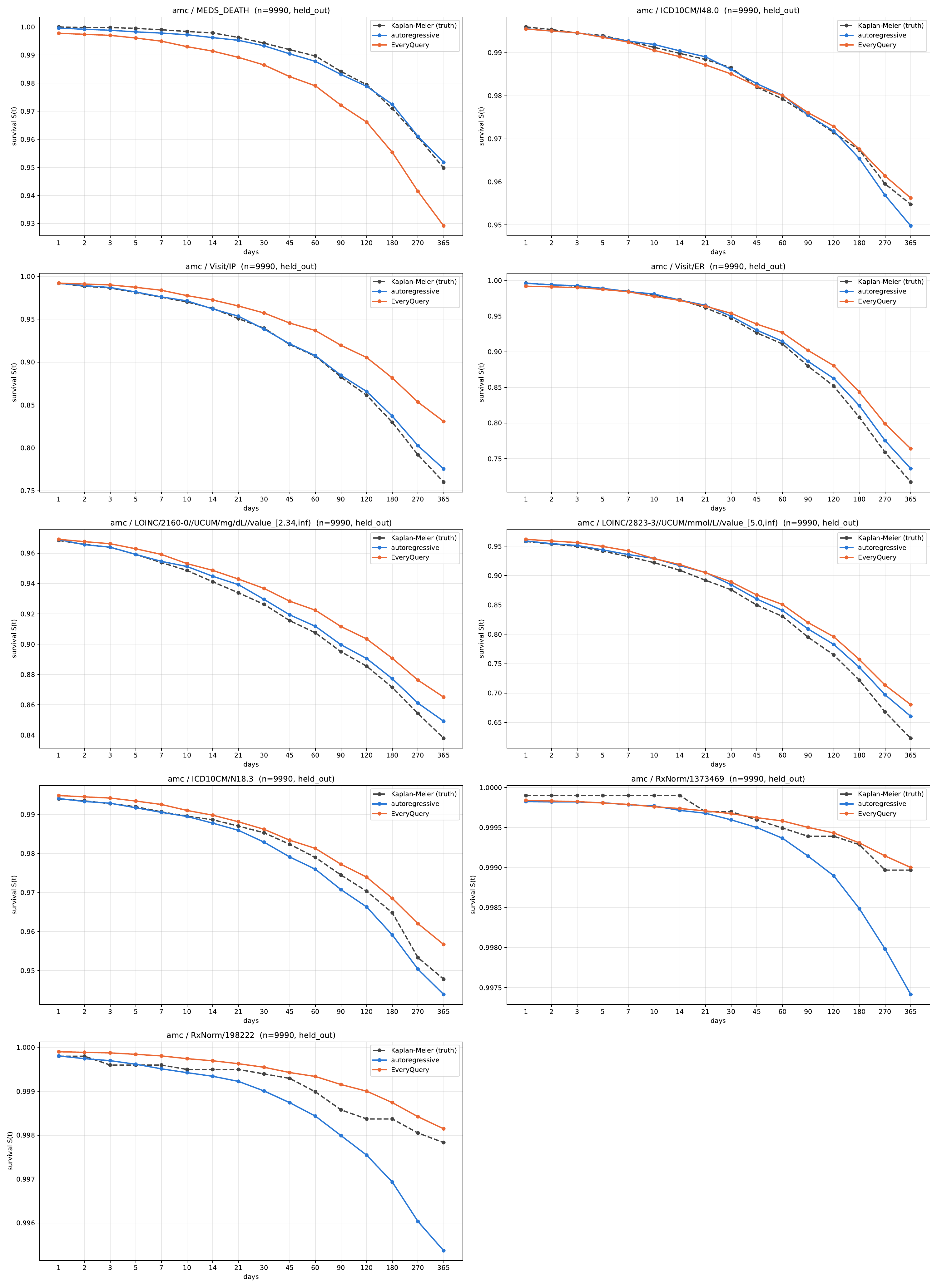}
\caption{\textbf{Survival curves on AMC.} Each panel shows one target code (Table~\ref{tab:tte_codes}), comparing the Kaplan--Meier estimate on held-out patients (black, dashed) with the survival curves estimated by the autoregressive baseline (blue) and \model (orange) at each prediction horizon in days.}
    \label{fig:amc-survival-curves}
\end{figure}

Figures~\ref{fig:mimic-survival-curves}, \ref{fig:nwicu-survival-curves} and \ref{fig:amc-survival-curves} compare the survival curves estimated by \model and the autoregressive baseline with the Kaplan--Meier estimate for every target code on MIMIC-IV, NWICU and AMC.

\clearpage 

\section{Results on Randomly Sampled Tasks}
\label{app:random}

\begin{figure}[H]
    \centering
    \includegraphics[width=\textwidth]{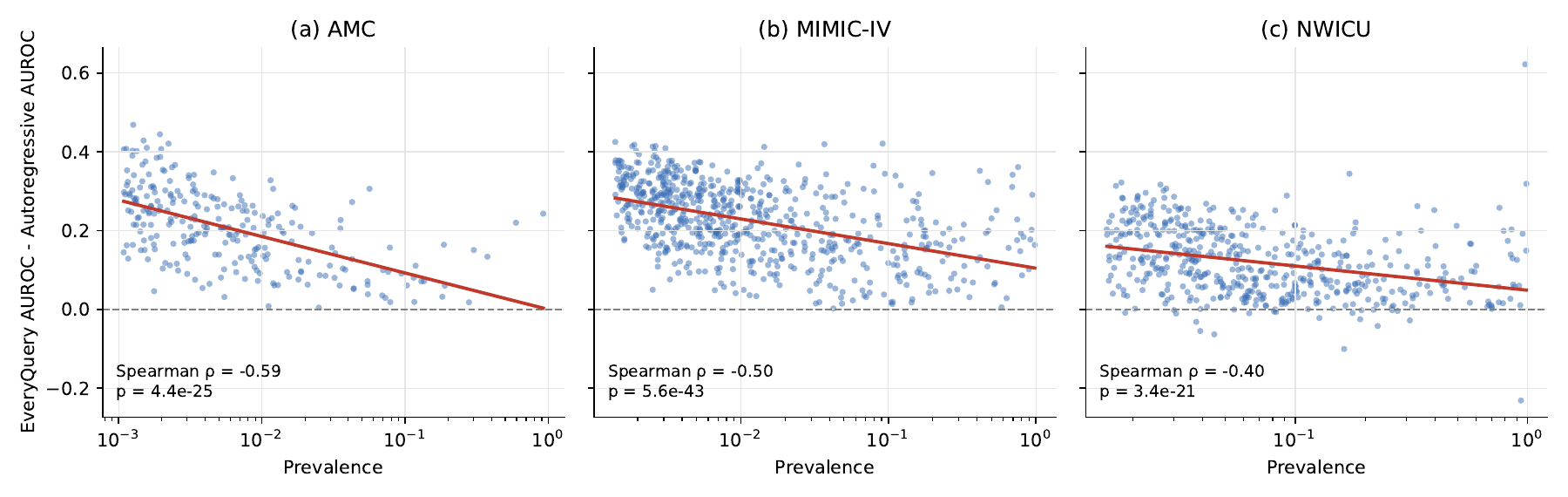}
\caption{\textbf{On randomly sampled tasks, \model's advantage over the autoregressive baseline grows as events become rarer.} Each point is one randomly sampled task (\S\ref{sec:random-tasks}), plotted by prevalence (log scale) against $\Delta$AUROC (\model minus autoregressive). Red lines show linear fits, and each panel reports the Spearman correlation between prevalence and $\Delta$AUROC.}
    \label{fig:random-task-prev}
\end{figure}

\end{document}